\documentclass{article}

\usepackage[preprint,nonatbib]{neurips_2020}

\usepackage[utf8]{inputenc} 
\usepackage[T1]{fontenc}    
\usepackage{hyperref}       
\usepackage{nicefrac}       
\usepackage{microtype}
\usepackage{graphicx}
\usepackage{subfigure}
\usepackage{booktabs} 
\usepackage{amsmath}
\usepackage{amssymb}
\usepackage{bbm}
\usepackage{hyperref}
\DeclareMathOperator*{\argmin}{arg\,min} 

\title{\emph{RelEx}: A Model-Agnostic Relational Model Explainer}

\author{
  Yue Zhang \\
  Department of Computer Science, SUNY Binghamton \\
  \texttt{yzhan202@binghamton.edu} \\
  \AND
  David Defazio  \\
  SUNY Binghamton \\
  \texttt{ddefazi1@binghamton.edu } \\
  \And
  Arti Ramesh \\
  SUNY Binghamton \\
  \texttt{artir@binghamton.edu} \\
}

\begin{document}

\maketitle

\begin{abstract}

In recent years, considerable progress has been made on improving the interpretability of machine learning models. This is essential, as complex deep learning models with millions of parameters produce state of the art results, but it can be nearly impossible to explain their predictions. While various explainability techniques have achieved impressive results, nearly all of them assume each data instance to be independent and identically distributed (iid). This excludes relational models, such as Statistical Relational Learning (SRL), and the recently popular Graph Neural Networks (GNNs), resulting in few options to explain them. While there does exist one work on explaining GNNs, GNN-Explainer, they assume access to the gradients of the model to learn explanations, which is restrictive in terms of its applicability across non-differentiable relational models and practicality. In this work, we develop \emph{RelEx}, a \textit{model-agnostic} relational explainer to explain black-box relational models with only access to the outputs of the black-box. \emph{RelEx} is able to explain any relational model, including SRL models and GNNs. We compare \emph{RelEx} to the state-of-the-art relational explainer, GNN-Explainer, and relational extensions of iid explanation models and show that \emph{RelEx} achieves comparable or better performance, while remaining model-agnostic. 


\end{abstract}

\section{Introduction}

In the last decade, significant attention has been directed toward accurately modeling non-Euclidean, graph-structured data. Relational Models include Statistical relational learning (SRL) methods \cite{srl2007}, stochastic blockmodels, and the more recently developed Graph Neural Networks (GNNs) \cite{scarselli2008graph}. These relational models can be applied for a variety of tasks dealing with structured data, e.g. molecule classification, knowledge graph completion, and recommendation systems. 

Along with relational models, progress has also been made in explaining the predictions of black-box models. We use the term \textit{black-box models} to refer to models whose predictions are not inherently interpretable. For example, the pixels that are instrumental in a prediction is not often apparent for state-of-the-art deep neural network models for many vision tasks \cite{goodfellow2014explaining}. This has led to the emergence of models aimed at explaining the predictions of complex underlying models, however most underlying approaches are designed to only work on independent and identically distributed (iid) data. There are mainly two groups of these iid model explainers. The first group finds important data points, which have high influence on learnt model behavior, including influence function (IF) \cite{koh2017understanding}, and representer points \cite{yeh2018representer}. The second group finds feature attributes that are most influential to the final model decision \cite{sundararajan2017axiomatic,smilkov2017smoothgrad,ribeiro2016should,ribeiro2018anchors}. However, explaining relational data and relational models is significantly more challenging as it involves learning the right relational structure around the node of interest that explains the prediction. There is limited existing work on explaining relational data and relational models, in fact, to our knowledge, there is only one such technique GNN-Explainer \cite{ying2019gnn}, which has been designed for explaining models that consider dependencies amongst the data samples. GNN-Explainer learns the most important neighbor nodes and links corresponding to why a GNN predicts some nodes as a specific class. This explanation is learned as masks over the adjacency and feature matrices, which are optimized by utilizing the gradients of the underlying GNN model. 

In this paper, we present \emph{RelEx}, a model-agnostic relational model explainer that learns relational explanations by treating the underlying model as a black-box model. We construct explanations first by learning a local differentiable approximation of the black-box model for some node of interest, trained over the perturbation space of this node. We then learn an interpretable mask over the local approximation. 

Specifically, our contributions are as follows: i) We develop \emph{RelEx}, which learns model-agnostic relational explanations for the task of node classification, with only access to the output prediction of the black-box model for a specific input. Hence, \emph{RelEx} can be applied to any relational model, from non-differentiable statistical relational models to various GNNs. ii) \emph{RelEx} can learn diverse explanations for each data instance by maximizing the cross-entropy between two learned relational explanations. This provides end users with the much-needed flexibility of choosing an explanation that is more appealing from a domain perspective, while remaining true to the underlying black-box model. iii) We perform experiments on both synthetic and real-world datasets, comparing our relational explanations to the correct ground-truth relational structures (we refer to them as right reasons \cite{ross2017right}). We demonstrate that our approach is comparable to or better than the state-of-the-art relational explainer, GNN-Explainer, and relational extensions of other state-of-the-art explainers in quantitative performance across all datasets, despite needing less information about the black-box model than these approaches. We also illustrate the capability of \emph{RelEx} to capture the core topological structures in the explanations in different classification tasks through qualitative results across all the datasets. 
	
Thus, \emph{RelEx} is model-agnostic and practically more feasible than existing approaches to explaining relational models. \emph{To the best of our knowledge, ours is the first general-purpose model-agnostic relational explainer.}


\section{Related Work}

In this work, we specifically focus on explaining two broad types of relational models: statistical relational learning, and graph neural networks. Since the primary focus of this work is explaining relational models, we only discuss the different relational models briefly before discussing explanation approaches.

Statistical relational learning (SRL) \cite{srl2007} is concerned with domain models that exhibit both uncertainty and complex, relational structure, where a user handcrafts first-order logic rules to capture dependencies and reasoning. For example, the collective rule: $\lambda$, \textit{Spouse(B,A) $\land$ Votes(A,C) $\to$ Votes(B,C)} captures the increased probability of a spouse to vote for the same candidate in an election, as determined by the target variable \textit{Votes} and observed variable \textit{Spouse} dependencies, and $\lambda$ is the rule weight. Hinge-Loss Markov Random Fields (HL-MRFs) \cite{bach2017hinge} is an example of an SRL model which uses declarative language Probabilistic Soft Logic (PSL) to express probabilistic logic rules. The input of HL-MRFs is a knowledge graph $(P, R, O)$, where $P$ is a set of entities, $R$ is a set of relations, and $O$ is a set of observed relational data of the format $r(h, t)$, where $h$ and $t$ are entities and $r$ is a binary relation between entities. Example of such data tuple is \textit{Spouse(Alice, Bob)}. For each relation $r$, we can represent our knowledge using adjacency matrix $A_r=\{0,1\}^{|P|\times|P|}$, such that its $(i,j)$ entry is 1 if and only if $r(P_i, P_j)$ is in the knowledge graph. We choose HL-MRFs as a representative SRL model to evaluate our relational explainer.

GNNs are another more recent technique for modeling relational data, with a variety of popular architectures \cite{kipf2017semi,xu2018how,velickovic2018graph}. Given an adjacency matrix defining the relations among nodes, and a feature matrix describing the attributes of each node, a GNN will learn low-dimensional node representations, similar to classical feed forward neural networks. Representations are initialized to default node features, and are updated in each layer through degree normalized aggregations of its neighbor's node representations. After training, these node representations capture the task relevant feature and structural information, and can then be used for a variety of machine learning tasks. In this paper, we evaluate \emph{RelEx} for explaining GNNs for the task of node classification. 


Post-hoc explainers such as LIME \cite{ribeiro2016should} have been developed to learn an interpretable local approximation of a black-box, to explain single instances. Our relational explainer \emph{RelEx} is motivated by this approach. Anchors \cite{ribeiro2018anchors} were subsequently developed to make clear where LIME explanations apply. These approaches are model-agnostic, meaning they work regardless of how the black-box model works (so long as it's a model on iid data). Other explanation techniques involve using input gradients to learn feature importances. This includes \textit{SmoothGrad}\cite{smilkov2017smoothgrad}, and \textit{Integrate Gradient}\cite{sundararajan2017axiomatic}, where the basic idea behind them is to learn the saliency map by calculating gradient on input $\partial f(x)/\partial x$, however, they are designed specifically for image data. 


There has been limited effort to improve the interpretability of deep learning based methods on graphs. Graph Attention Networks is an architecture which learns attention weights on each edge, which can be interpreted as an importance score \cite{velickovic2018graph}. Another approach simplifies Graph Convolutional Networks (GCNs) by removing the nonlinear function applied to each layer \cite{pmlr-v97-wu19e}. It is shown that for many cases, performance is the same while the underlying classifier is equivalent to logistic regression and thus more interpretable. Another work attempts to disentangle node representations by capturing the latent factors in their neighborhoods \cite{ma2019disentangled}. While this can improve robustness and interpretability, GNN-Explainer \cite{ying2019gnn} is the only post-hoc approach to provide explanations for particular node predictions. 

Our approach addresses the following caveats in existing work. First, our approach only needs access to the output predictions of the black-box model, not the gradients. Thus, in contrast to GNN-Explainer, our approach is capable of explaining any relational model, including non-differentiable HL-MRFs. Our approach also shines from a practical usability perspective, as some popular developed models \cite{Fey/Lenssen/2019} use indices of existing edges rather than adjacency matrices as input. 

\vspace{-0.0cm}
\section{\emph{RelEx}: Learning-based Relational Explainer}
\begin{figure*}[t]
\centering
\includegraphics[width=0.9\textwidth]{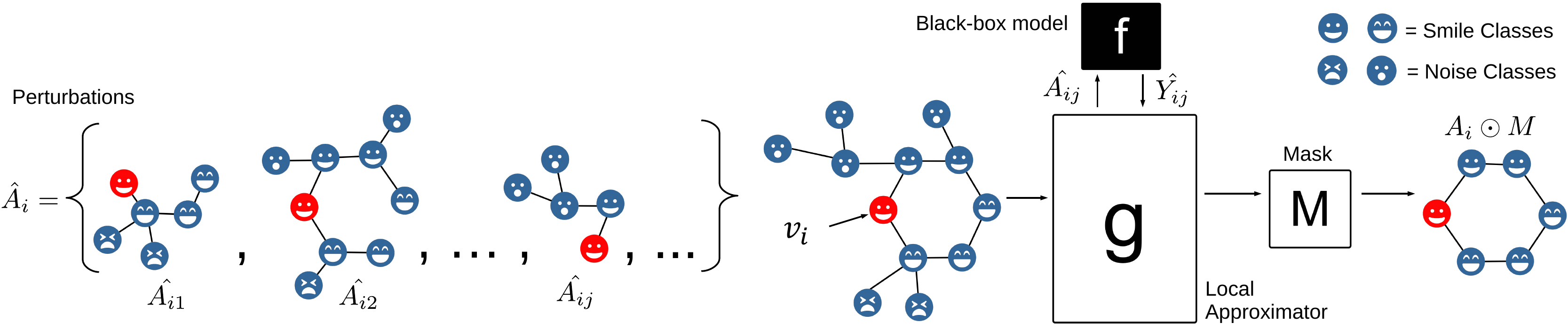}
\caption{\emph{RelEx} architecture showing the different components and the notations associated with them.} 
\label{fig:arch}
\end{figure*}

\vspace{-0.0cm}
In this section, we develop our model-agnostic, learning-based relational explainer, \emph{RelEx}, for explaining predictions of relational models. \emph{RelEx} learns which nodes and edges in the neighborhood of the node of interest are most influential in the black-box prediction. The output of \emph{RelEx} is a neighborhood relational structure that is instrumental in the black-box prediction. Figure \ref{fig:arch} gives the overall architecture of \emph{RelEx} and identifies the different components and their notations, which we use in the equations in this paper. 


\subsection{\emph{RelEx} Problem Formulation}

Let node $v_i$ be the node we want to explain. We denote the n-hop neighborhood of $v_i$ (i.e., the computation graph) by the adjacency matrix $A_i \in \{0,1\}^{P_i\times P_i}$, which contains a total of $P_i$ nodes and $E_i$ edges. The features of these neighborhood nodes are stored in the feature matrix $X_i$. We want to learn an explanation, where given some black-box relational model $f$, we learn the most salient nodes and links in $A_i$ that are instrumental in $f$'s prediction of $v_i$. We refer to the predicted class of $v_i$ as $Y_i$. $Y_i$ is represented as a one-hot vector. 


After defining the problem setting, we can see that the problem of learning a relational explanation for node $v_i$ entails selecting nodes and edges from $A_i$, the computation graph of $v_i$. Hence, to select the salient graph structures, we learn a sparse mask over $A_i$. Our relational explainer \textit{RelEx} is denoted by $\Phi(f, A_{i}) \in \{0,1\}^{P_i\times P_i}$, which is a sparse adjacency matrix consisting of the nodes and edges crucial to the explanation. 

$\Phi(f, A_{i})$ is learned by optimizing for the below loss function,
\begin{align}
\argmin_M J &= \textit{L}(Y_i, f(\Phi, X_i)) + \Psi(M) =  \textit{L}(Y_i, f(A_i\odot M, X_i)) + \Psi(M)
\label{eqn:loss}
\end{align}
where the loss function \textit{L} is any distance measure between our class of interest $Y_i$ and the probability distribution over classes predicted by the underlying black-box model $f$ on the de-noised computation graph. Common choices for \textit{L} are negative log-likelihood and \textit{KL}-divergence. We represent the explanation as $\Phi=A_i\odot M$, where $\odot$ denotes element-wise multiplication, and $M \in \{0,1\}^{P_i\times P_i}$ is the mask we are optimizing for. $\Psi$ is a sparseness measure on the mask $M$, which could be $L_1$ norm or a group sparseness measure like $L_{21}$ norm \cite{he2016adaptive}. We discuss $\Psi$ with greater detail in the sections to follow. As finding the optimal solution to Equation \ref{eqn:loss} is a combinatorial problem, a brute-force approach has the time complexity $\mathcal{O}(2^{E_i})$. There are a variety of heuristic search based solutions to this optimization problem, including multi-armed bandit \cite{ribeiro2018anchors}, reinforcement learning methods \cite{zhang2019learning}. 

\subsection{\emph{RelEx} Architecture}

Here, we present \emph{RelEx}, a more effective and efficient learning-based solution for explaining relational models. We first provide a general overview of our approach and then expand on the different components in the following paragraphs. To design  \emph{RelEx}, our first goal is to learn a local approximator of $f$ at the node $v_i$ whose prediction we are interested in explaining. We call this approximator $g(\hat{A_i})$, where $\hat{A_i}$ is a perturbation on the computation graph $A_i$ of $v_i$. Since $f$ is a relational graph model, we choose $g$ to be a naive Graph Convolutional Network (GCN), owing to its powerful fitting and representation ability, while simultaneously being easy to use. Specifically, we use a residual architecture \cite{dehmamy2019understanding}, where we concatenate the output of every GCN layer to the final output of the network. Residual architecture increases the representation power of GCN in learning graph topology without stacking more layers or adding more parameters. 

To learn $g$, we first follow a sampling strategy to get perturbed computation graphs $\hat{A_i}$. We then query $f$ on all samples of $\hat{A_i}$ to get dataset $S=\{\hat{A_{ij}}, \hat{Y_{ij}}\}$, where $j=1, 2, ..., n$, $n$ is the number of samples, and $\hat{Y_{ij}}=f(\hat{A_{ij}}, X_i)$. In contrast to LIME \cite{ribeiro2016should}, we do not require our local approximator to be interpretable itself as we learn a sparse explanation mask after learning the local approximator. This allows us to broaden the scope and complexity of the local approximator, thus, achieving the dual goals of expressibility and interpretability, whereas other existing models typically use simple models as the local approximators \cite{ribeiro2016should}. Hence, \emph{RelEx} is able to explain any black-box model as long as the local approximator is: i) locally faithful (this corresponds to how the model behaves in the vicinity of the instance being predicted), and ii) differentiable on the input adjacency matrix.


The modified objective function using the local approximator \textit{g} instead of the black-box model \textit{f} is given by,
\begin{align}
\argmin_M \hat{J} = L(Y_i, g(A_i\odot M)) + \Psi(M)
\end{align}
where we replace $f(\Phi, X_i)$ by $g(A_i\odot M)$. Below, we discuss more details on the different components in our relational explainer: i) sampling strategy used to create the perturbations, ii) functions for learning the sparse mask, iii) regularization, and iv) diverse explanations. 

\textbf{Sampling Strategy} \hspace{0.1cm}
We adopt a modified breadth first search (BFS) sampling strategy, starting from node $v_i$, where each connected edge has some fixed probability of being selected. We know that the nodes that are disjoint from $v_i$ will not affect a node's embedding or prediction, so we have no probability of selecting these nodes using BFS. Any node in $A_i$ has a chance to be sampled, so long as it is connected to an already selected node. We choose BFS as it encourages closer nodes to be selected more frequently. By sampling using BFS, we also ensure a higher value of variance amongst the farthest nodes in our samples. Thus, employing the BFS sampling strategy ensures that the closer nodes are to $v_i$, the higher their influence on the black-box's prediction of $v_i$ and we want our samples to be ``close" to $A_i$ in order for $g$ to be a \textit{local} approximation of $f$ at $v_i$. We do not select nodes outside of $A_i$, as they have no effect on $f$'s prediction of $v_i$. After each iteration of sampling, we get one connected perturbed subgraph. We construct the dataset $S$ by perturbing $A_i$ for multiple iterations. We learn the local approximator $g$ by training it on $S$.

\textbf{$\emph{RelEx}_\textnormal{Sigmoid}$: \emph{RelEx} with Sigmoid Mask} \hspace{0.1cm}
To get the mask $M$, we set $M=\sigma(W)$, where $\sigma$ is the sigmoid function, and $W$ is the parameter we need to learn. $M_{p,q} \in [0,1]$, which means we learn a soft mask, and each element in the mask represents the importance of the corresponding edge. The objective function with the sigmoid mask is given by, $\argmin_W \hat{J} = Loss(Y_i, g(A_i\odot \sigma(W))) + \Psi(\sigma(W))$. This optimization problem can be solved by gradient descent, where $W=W-\alpha*\frac{\partial J}{\partial W}$, and $\alpha$ is the learning rate. 

\textbf{$\emph{RelEx}_\textnormal{Gumbel}$: \emph{RelEx} with Gumbel-Softmax Mask} \hspace{0.1cm}
We introduce another mask based on Gumbel-softmax \cite{jang2016categorical}, $M=\text{Gumbel-softmax}(W)$, where $W$ is our parameter to be learned, and $M_{p,q} \in \{0,1\}$. This directly gives us a set of edges and nodes, unlike the Sigmoid mask where we learn a soft mask and then use a threshold. Choosing a threshold could be difficult, because we need to use the right reason as reference. And sometimes it is challenging to find the optimal threshold as learned soft values are close to each other. Gumbel-softmax is a continuous distribution on the simplex that can approximate categorical samples, and whose parameter gradients can be easily computed via the reparameterization trick. We have an end-to-end relational structures learning framework by using Gumbel-Softmax based method. 

\textbf{Regularization} \hspace{0.1cm}
We incorporate regularization measures in our objective to ensure that the learned mask remains sparse and has better interpretability. We consider two regularization functions. First, we incorporate sparseness on the edges using $L_1$ norm. Second, we incorporate $L_{2,1}$ norm \cite{he2016adaptive} on nodes and edges, given by $||M||_{2,1}=\sum_{i=1}^{P}\sqrt{\sum_{j=1}^{P}M_{ij}^2}$. $L_{2,1}$ is a group sparseness measure, here we treat each row of adjacency matrix as one group. Thus, we pursue sparseness on both edges and nodes. 

\textbf{Diverse Explanations} \hspace{0.1cm} There can be multiple explanations that exist for one prediction, but some are closer to the ``right'' reason \cite{ross2017right}. We encourage diversity by learning different masks and maximizing the cross-entropy loss between any two masks using Equation \ref{eqn:diversity}.
\begin{align}
\hat{J}^{(t)} &= L(Y_i, g(A_i\odot M^{(t)}))+ \Psi(M^{(t)}) -\alpha_H*(H(M^{(t)}, M^{(1)})+ ... + H(M^{(t)}, M^{(t-1)}))
\label{eqn:diversity}
\end{align}
where $M^{(t)}$ is the current mask to be learned, $M^{(1)}$, $M^{(2)}$, ..., $M^{(t-1)}$ are previously learned masks, $t \in [1,T]$, and $H(,)$ is cross entropy loss between two masks, $\alpha_H$ is the weight of cross entropy loss. 

Learning diverse explanations can increase the users' trust of black-box models \cite{mothilal2020explaining}. In our experiments in Section \ref{sec:diverseExperiment}, we show that our approach learns diverse masks, where at least one of them corresponds to the right reason.



\section{Evaluation Methods}
\subsection{Comparison with State-of-the-art Explainers}
\label{sec:relanchors}
We make appropriate modifications to Anchors \cite{ribeiro2018anchors} and Saliency Map to adapt them for the relational setting.

\textbf{Relational Anchors} \hspace{0.1cm} Anchor \cite{ribeiro2018anchors} explanations are constructed by selecting the features of some instance $x$ we want to explain that maximizes precision. This is calculated by perturbing $x$ over all features except the anchor features, which are held constant. Precision is the proportion of the sample's labels that do not change due to perturbations. A high precision anchor implies that the anchor features are most important to the prediction, because perturbing the other features has little or no effect. Approximating precision is difficult, as it requires many expensive calls to $f$. Therefore, anchor construction is formulated as an instance of pure-exploration multi-armed bandit. To adapt Anchors to our setting, we consider the anchor features for relational explanations to be graph edges instead of node features. We define a threshold $\delta$, which we vary based on how much the predictions of $f$ vary based on our perturbations. We calculate precision as, $\text{Precision}(Anc) = \mathbbm{E}_{D(z|Anc)}\bigg[\mathbbm{1} \Big[ 1 - f(z)_{c} < \delta \Big] \bigg] $, where $Anc$ is the set of anchor edges, $z$ is a perturbed sample, and $f$ is the black box. Our perturbed samples $z$ have all edges in $Anc$, and at most all edges in the computation graph of $v_i$. Samples are generated via breadth-first search, similarly to \emph{RelEx}, and $c$ is the predicted class we want to explain. 

\textbf{Saliency Map} \hspace{0.1cm} Since our approach involves learning an edge importance mask, we also compare our approach to Saliency Map, which is used in computer vision to learn the spatial support of a given target class in an image. To adapt Saliency Map to the relational setting, we first calculate the gradient of black-box model's loss function with respect to the adjacency matrix $A_i$, and then normalize the gradient values to values between $0$ to $1$. We use these values as the learned explanation.


\subsection{Relational Explanation Evaluation Metrics}

\textbf{Area Under the ROC Curve} \hspace{0.1cm} We report area under the receiver operating characteristics curve (AUC-ROC) by capturing the deviation of the explanation from the ground-truth right reasons. AUC-ROC is calculated between the relational explanation and ground truth right reason structure. In many cases, we know the right structural reasons associated with a prediction as prior domain knowledge \cite{ross2017right}. For example, molecules or proteins have their own specific and identifiable structures. We can then evaluate our explanations by comparing them to these already known right reasons. 

\textbf{Infidelity Scores} \hspace{0.1cm}
Since it is hard to isolate errors that stem from the underlying black-box model from the errors that stem from the explainer, we consider the following quantitative measure known as \textit{infidelity }\cite{yeh2019fidelity}, $\text{Infidelity}(\Phi,f,A_i) = E_{I\sim\mu_I}[(\textrm{sum}(I\odot\Phi(f,A_i))-(f(A_i)-f(A_i-I)))^2]$, where $I$ represents significant perturbations around the node $N_i$, $\mu_I$ gives the distribution of perturbation $I$, $\hat{A_i}=A_i-I$ represents the perturbed adjacency graph, and $\Phi$ is our explainer. Infidelity measures the goodness of an explanation by quantifying the degree to which it captures how the predictor function itself changes in response to a significant perturbation.

\section{Experiments}

\subsection{Experiments on Synthetic Relational Datasets with GNNs as the Black-box Model}
We construct two kinds of node classification datasets: i) \textsc{Tree-Grid}, in which we use a binary tree with fixed height as the basic structure, and then we connect multiple grid structures to the tree, by randomly adding noisy links between nodes in a grid (referred to as grid nodes) and a tree (tree nodes), and ii) \textsc{Tree-BA}, we again use a binary tree as the basic structure (tree nodes), and then we connect multiple Barabasi Albert (BA) structures (BA nodes) to the tree, by randomly adding noise in the form of links between BA nodes and tree nodes. The prediction problem involves predicting the correct class to which each node belongs using the neighborhood topological structures of the node.

%

\begin{table}
	\begin{centering}
	\scriptsize
	\caption{AUC-ROC and infidelity for \textsc{Tree-Grid} synthetic dataset.}
	\begin{tabular}{p{1.8cm}p{1.8cm}p{2.0cm}p{1.8cm}p{1.8cm}p{1.8cm}}
	\toprule
 	Explainer & Saliency Map & Relational Anchors & GNN-Explainer & \emph{RelEx$_\text{Sigmoid}$} & \emph{RelEx$_\text{Gumbel}$} \\
 	\midrule
	AUC-ROC & 0.4352 & 0.5069 & 0.5666 & 0.5470 & \textbf{0.5873} \\
 	Infidelity & 0.1199 & 0.1110 & 0.0885 & 0.0893 & \textbf{0.0884} \\
 	\bottomrule
	\end{tabular}
    \label{table:GNN_grid_results}
	\end{centering}

	\begin{centering}
	\scriptsize
	\caption{AUC-ROC and infidelity for \textsc{Tree-BA} synthetic dataset.}
	\begin{tabular}{p{1.8cm}p{1.8cm}p{2.0cm}p{1.8cm}p{1.8cm}p{1.8cm}}
	\toprule
 	 	Explainer & Saliency Map & Relational Anchors & GNN-Explainer & \emph{RelEx$_\text{Sigmoid}$} & \emph{RelEx$_\text{Gumbel}$} \\
 	\midrule
	AUC-ROC 	& 0.1205 & 0.6871 & 0.8431 & 0.8261 & \textbf{0.8672} \\
	Infidelity & 0.1317 & 0.0754 & 0.0782 & 0.0794 & \textbf{0.0735} \\
 	\bottomrule
	\end{tabular}
    \label{table:GNN_BA_results}
	\end{centering}
\end{table}

We train a 3-layer GCN as the black-box on both the datasets, individually. We show quantitative results on \textsc{Tree-Grid} dataset in Table \ref{table:GNN_grid_results} and quantitative results on \textsc{Tree-BA} dataset in Table \ref{table:GNN_BA_results}. Since both \textsc{Tree-Grid} and \textsc{Tree-BA} datasets are synthetically generated, we know the ground truth right reason structure and we use that to calculate the deviation of learned relational explanation from the right reason. From the AUC-ROC and infidelity results, we can see that \emph{RelEx$_\text{Gumbel}$} has the best performance on both measures. We also note that Saliency Map fails to perform as well as other models, as the former is the only model that is not specifically tailored for relational models. This further ascertains that explainers designed for traditional iid models do not seamlessly work for relational models and we need explainers that are designed specifically for relational models.

\subsection{Experiments on Synthetic Relational Dataset with HL-MRFs as the Black-box Model}
We construct a three-class graph dataset, \textsc{Tree-Grid-BA}. We generate multiple tree, grid, and Barabasi Albert motifs, and randomly add noisy links among them to construct this graph dataset. For HL-MRF models, we design collective first-order logic rules in table \ref{table:rules}. To train a PSL model, we randomly select half the nodes as observation, which are used as seed nodes.


\begin{table}%
\parbox{0.5\columnwidth}{
	\caption{3-hop PSL collective Rules}
	\scriptsize
	\begin{tabular}{p{6.5cm}}
	\toprule
	\textbf{PSL Collective Rules}\\
	\midrule
	Node: $A$, $B$, $C$, $D$; Target Class: $cat$\\
	\midrule
	$\lambda_1$: \textit{HasCat(A,cat) $\land$ Link(A,B) $\to$ HasCat(B,cat)} \\
    $\lambda_2$: \textit{HasCat(A,cat) $\land$ Link(A,B) $\land$ Link(B,C) $\to$ HasCat(C, cat)} \\
	$\lambda_3$: \textit{HasCat(A,cat) $\land$ Link(A,B) $\land$ Link(B,C) $\land$ Link(C,D) $\to$ HasCat(D, cat)} \\
	\bottomrule
	\end{tabular}
    \label{table:rules}
	}  
\qquad
\begin{minipage}[c]{0.4\textwidth}%
    \scriptsize
		\caption{AUC-ROC and infidelity for TREE-GRID-BA synthetic dataset.}
	\begin{tabular}{p{1.2cm}p{1.0cm}p{1.0cm}p{1.0cm}}
	\toprule
 	Explainer & Relational Anchors & \emph{RelEx$_\text{Sigmoid}$} & \emph{RelEx$_\text{Gumbel}$} \\
 	\midrule
	AUC-ROC & {0.5221} & \textbf{0.7076} & 0.6284 \\
	Infidelity & 0.0396 & \textbf{0.0310} & 0.0320 \\
 	\bottomrule
	\end{tabular}
    \label{table:PSL_results}
  \end{minipage}
\end{table}

\begin{figure}[t]
\centering
\subfigure[\scriptsize{Tree Computation Graph}]{
\includegraphics[width=0.15\columnwidth]{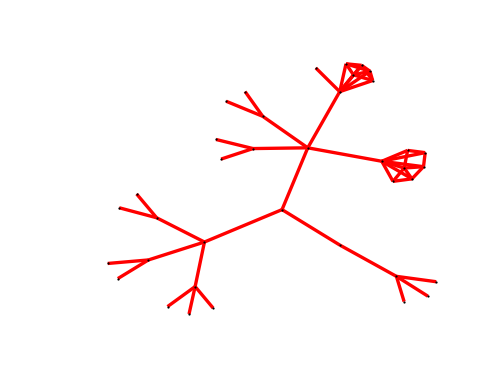}
\label{fig:PSL_comp_tree}
}
\centering
\subfigure[\scriptsize{Tree Right Reason}]{
\includegraphics[width=0.15\columnwidth]{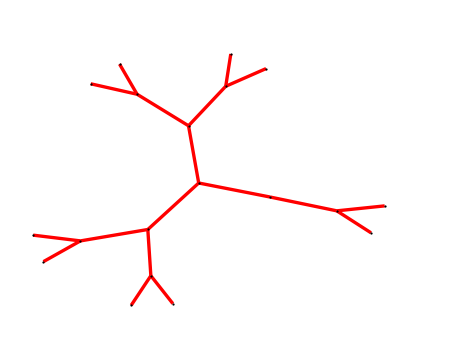}
\label{fig:PSL_rr_tree}
}
\centering
\subfigure[\scriptsize{Tree \emph{RelEx$_\text{Sigmoid}$}}]{
\includegraphics[width=0.15\columnwidth]{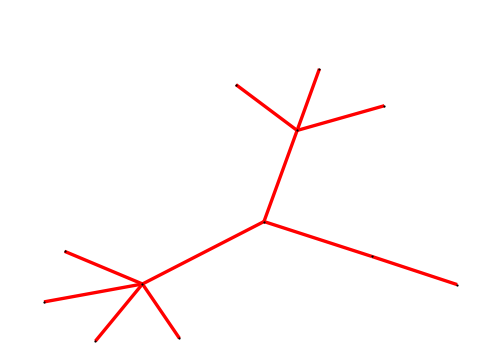}
\label{fig:PSL_sigmoid_tree}
}
\centering
\subfigure[\scriptsize{Tree \emph{RelEx$_\text{Gumbel}$}}]{
\includegraphics[width=0.15\columnwidth]{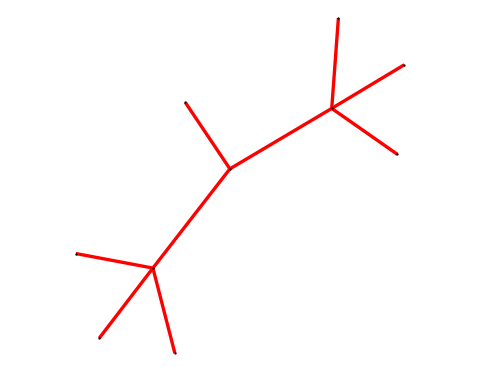}
\label{fig:PSL_gumbel_tree}
}
\centering
\subfigure[\scriptsize{Grid Computation Graph}]{
\includegraphics[width=0.15\columnwidth]{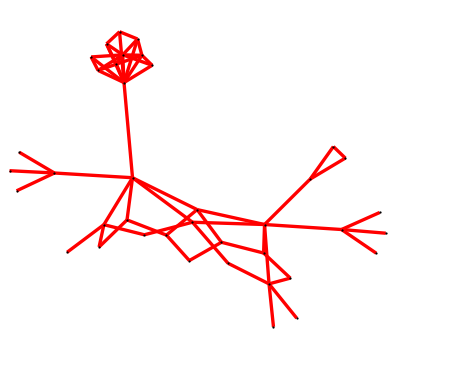}
\label{fig:PSL_comp_grid}
}
\centering
\subfigure[\scriptsize{Grid Right Reason}]{
\includegraphics[width=0.14\columnwidth]{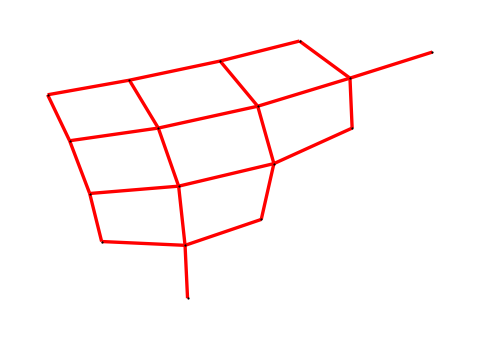}
\label{fig:PSL_rr_grid}
}
\centering
\subfigure[\scriptsize{Grid \emph{RelEx$_\text{Sigmoid}$}}]{
\includegraphics[width=0.15\columnwidth]{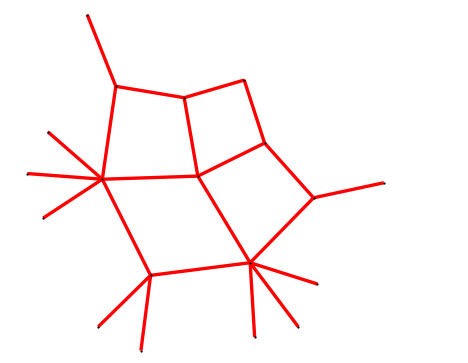}
\label{fig:PSL_sigmoid_grid}
}
\centering
\subfigure[\scriptsize{Grid \emph{RelEx$_\text{Gumbel}$}}]{
\includegraphics[width=0.15\columnwidth]{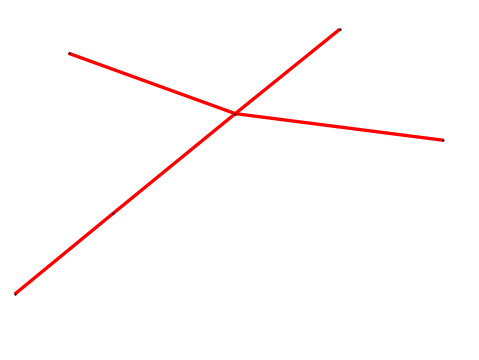}
\label{fig:PSL_gumbel_grid}
}
\centering
\subfigure[\scriptsize{BA Computation Graph}]{
\includegraphics[width=0.15\columnwidth]{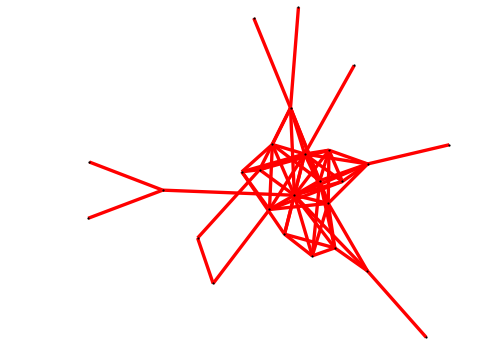}
\label{fig:PSL_comp_BA}
}
\centering
\subfigure[\scriptsize{BA Right Reason}]{
\includegraphics[width=0.15\columnwidth]{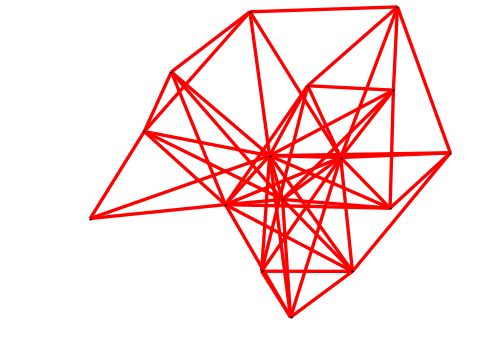}
\label{fig:PSL_rr_BA}
}
\centering
\subfigure[\scriptsize{BA \emph{RelEx$_\text{Sigmoid}$}}]{
\includegraphics[width=0.15\columnwidth]{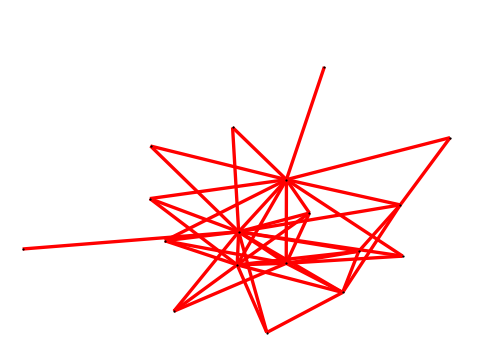}
\label{fig:PSL_sigmoid_BA}
}
\centering
\subfigure[\scriptsize{BA \emph{RelEx$_\text{Gumbel}$}}]{
\includegraphics[width=0.14\columnwidth]{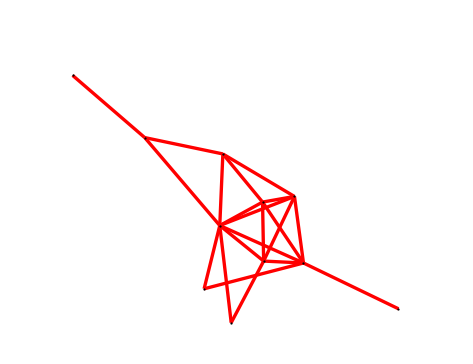}
\label{fig:PSL_gumbel_BA}
}
\caption{Explanations for tree, grid, \textit{BA} nodes.} 
\label{fig:PSL_syhthetic}
\end{figure}

Since GNN-Explainer and Saliency Map need access gradients, they cannot be applied to black-box HL-MRF models. Quantitative results are shown in Table \ref{table:PSL_results}, where we can see \emph{RelEx$_\text{Sigmoid}$} gets better results than \emph{RelEx$_\text{Gumbel}$}, as the HL-MRF model assigns different continuous values of importance to links around the node of interest, which are captured by the learned rule weights. Thus, \emph{RelEx$_\text{Sigmoid}$} successfully learns corresponding importance values for each link. This shows the competence of both our approaches \emph{RelEx$_\text{Sigmoid}$} and \emph{RelEx$_\text{Gumbel}$} across two different types of relational models. Figure \ref{fig:PSL_syhthetic} shows example explanation of a tree node, grid node, and BA node, respectively. We observe that the qualitative results are consistent with the quantitative results with \emph{RelEx$_\text{Sigmoid}$}, obtaining relational explanations that are closer to the actual right reason. We also can see that the \emph{RelEx$_\text{Sigmoid}$} model is able to glean the core topological structure that explains the prediction. 

\subsection{Experiments on Molecule Dataset with GNNs as the Black-box Model}

To demonstrate the applicability of our approach on a real-world dataset, we conduct experiments on MUTAG \cite{doi:10.1021/jm00106a046}, a well-known benchmark graph classification dataset. It consists of 188 mutagenic aromatic and heteroaromatic nitro compounds with 7 different kinds of atoms, including carbon, nitrogen and oxygen, etc. We have prior domain knowledge that carbon atoms have ring structures, which represent mutagenic aromatics in chemistry; nitrogen atoms and oxygen atoms combine to form the $\text{-NO}_2$ structure and nitrogen atoms also could exist either in pentagonal or hexagonal structures with other carbon atoms.


\begin{table}
	\begin{centering}
	\scriptsize
		\caption{Infidelity on MUTAG dataset.}
	\begin{tabular}{p{1.8cm}p{1.8cm}p{2.0cm}p{1.8cm}p{1.8cm}p{1.8cm}}
	\toprule
 	Explainer & Saliency Map & Relational Anchors & GNN-Explainer & \emph{RelEx$_\text{Sigmoid}$} & \emph{RelEx$_\text{Gumbel}$} \\
 	\midrule
	Infidelity & 0.05879 & 0.06008 & \textbf{0.05557} & 0.05659 & \textbf{0.05573} \\
 	\bottomrule
	\end{tabular}
    \label{table:MUTAG_results}
	\end{centering}
\end{table}

\begin{figure}[t]
\centering
\subfigure[\scriptsize{Molecule}]{
\includegraphics[width=0.14\columnwidth]{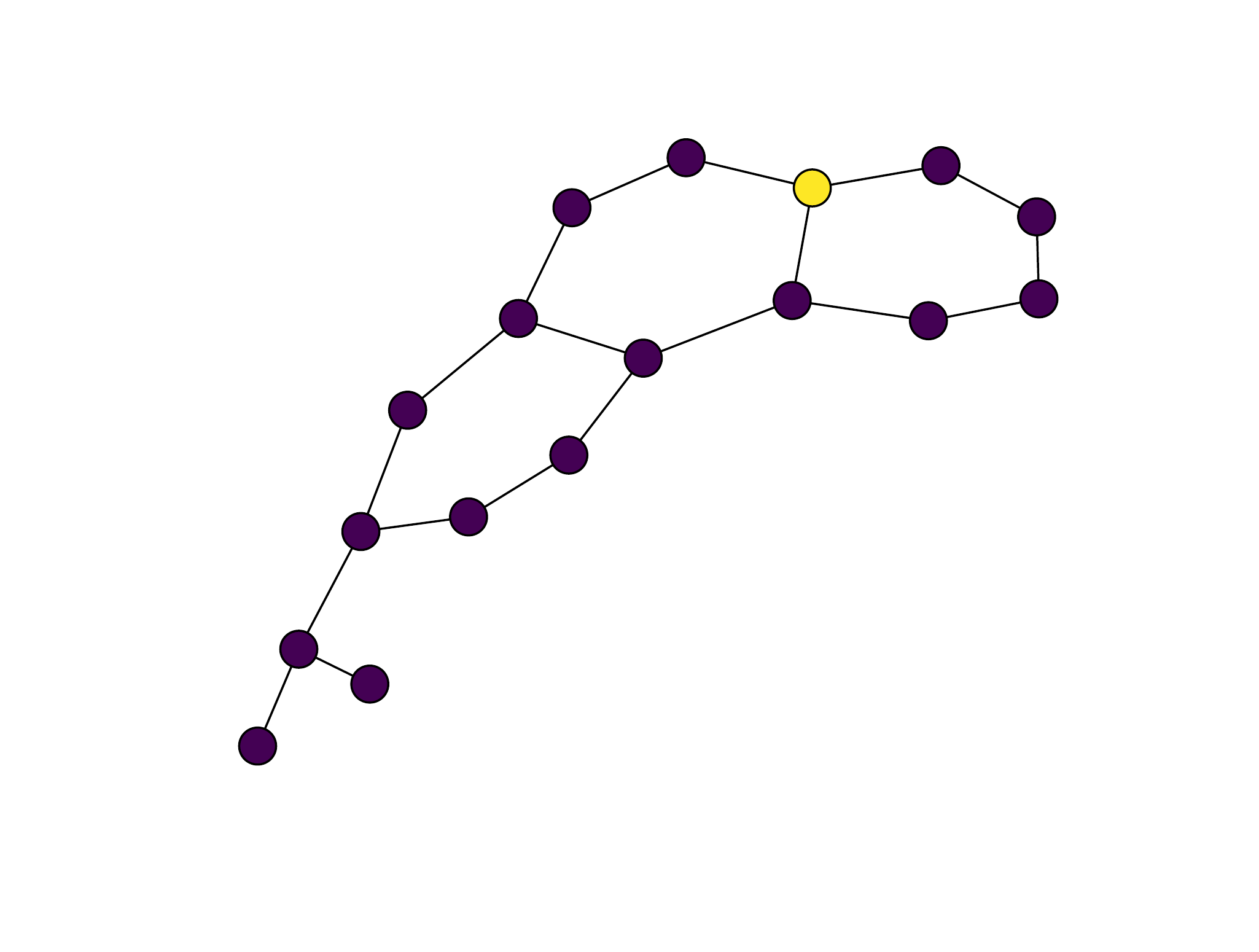}
\label{fig:molecule_m1}
}
\centering
\subfigure[\scriptsize{Right Reason}]{
\includegraphics[width=0.14\columnwidth]{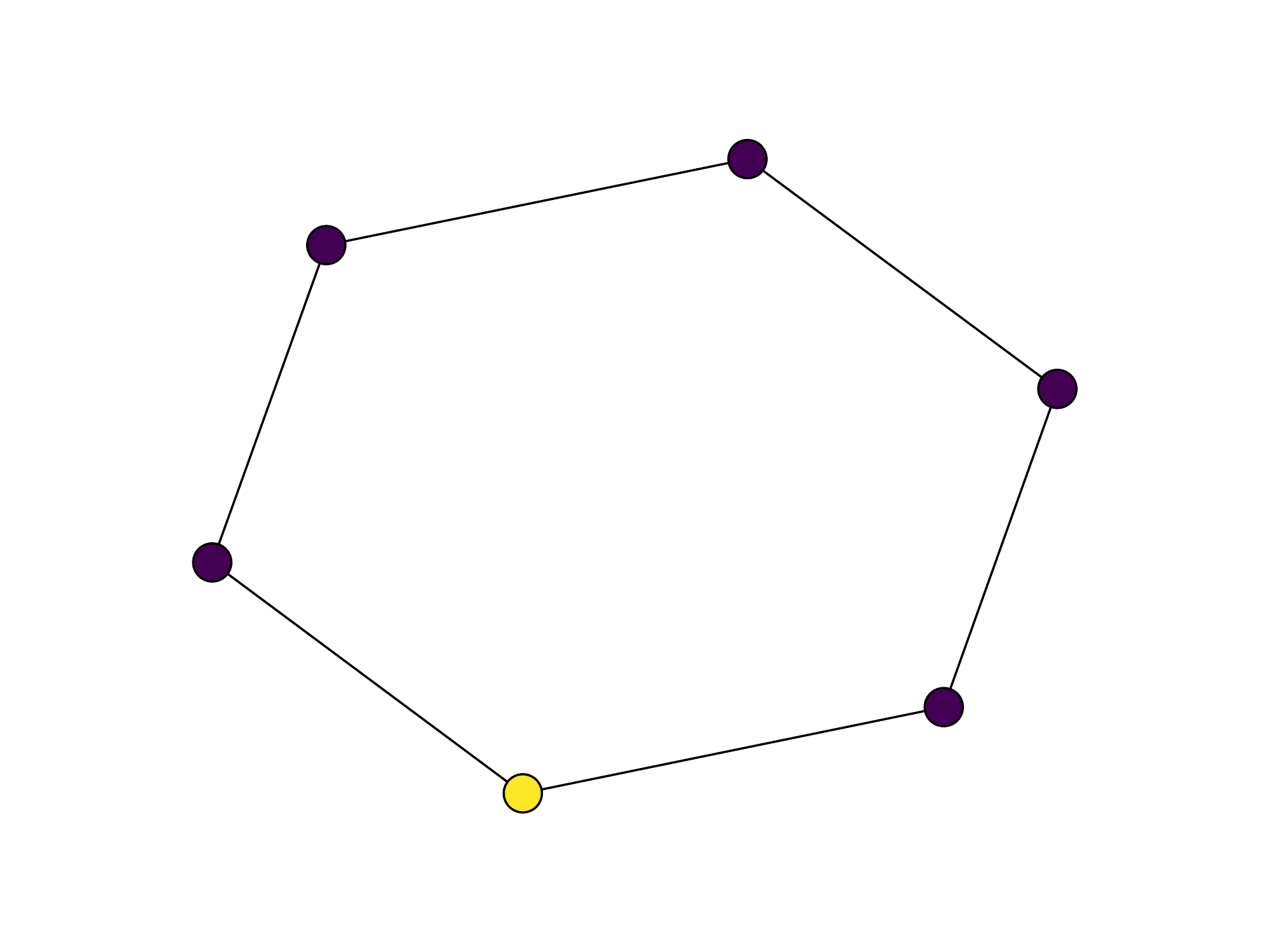}
\label{fig:rr_m3}
}
\centering
\subfigure[\scriptsize{Relational Anchors}]{
\includegraphics[width=0.14\columnwidth]{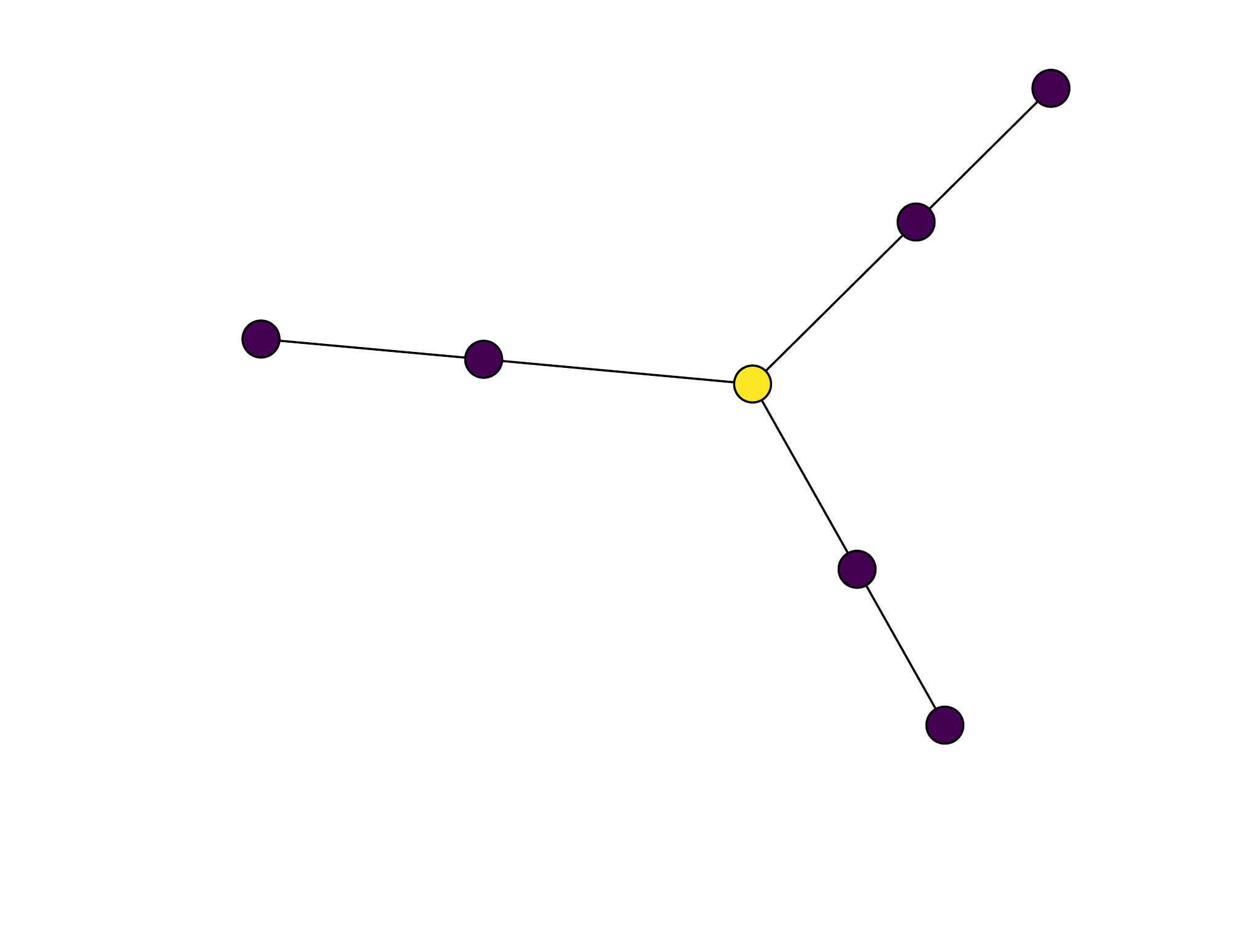}
\label{fig:anchor_m3}
}
\centering
\subfigure[\scriptsize{GNN-Explainer}]{
\includegraphics[width=0.14\columnwidth]{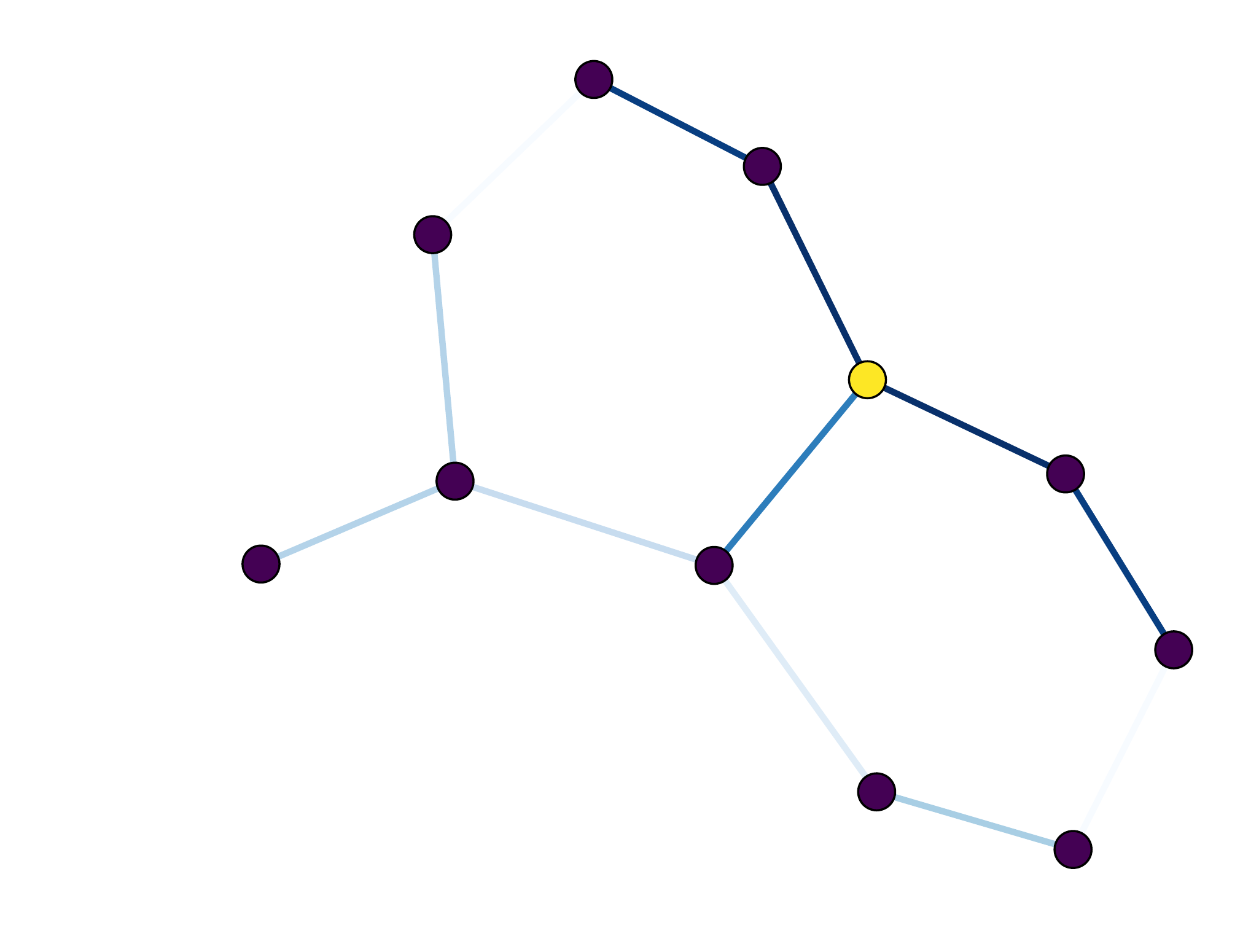}
\label{fig:gnnexplainer_m2}
}
\centering
\subfigure[\scriptsize{\emph{RelEx$_\text{Sigmoid}$} }]{
\includegraphics[width=0.14\columnwidth]{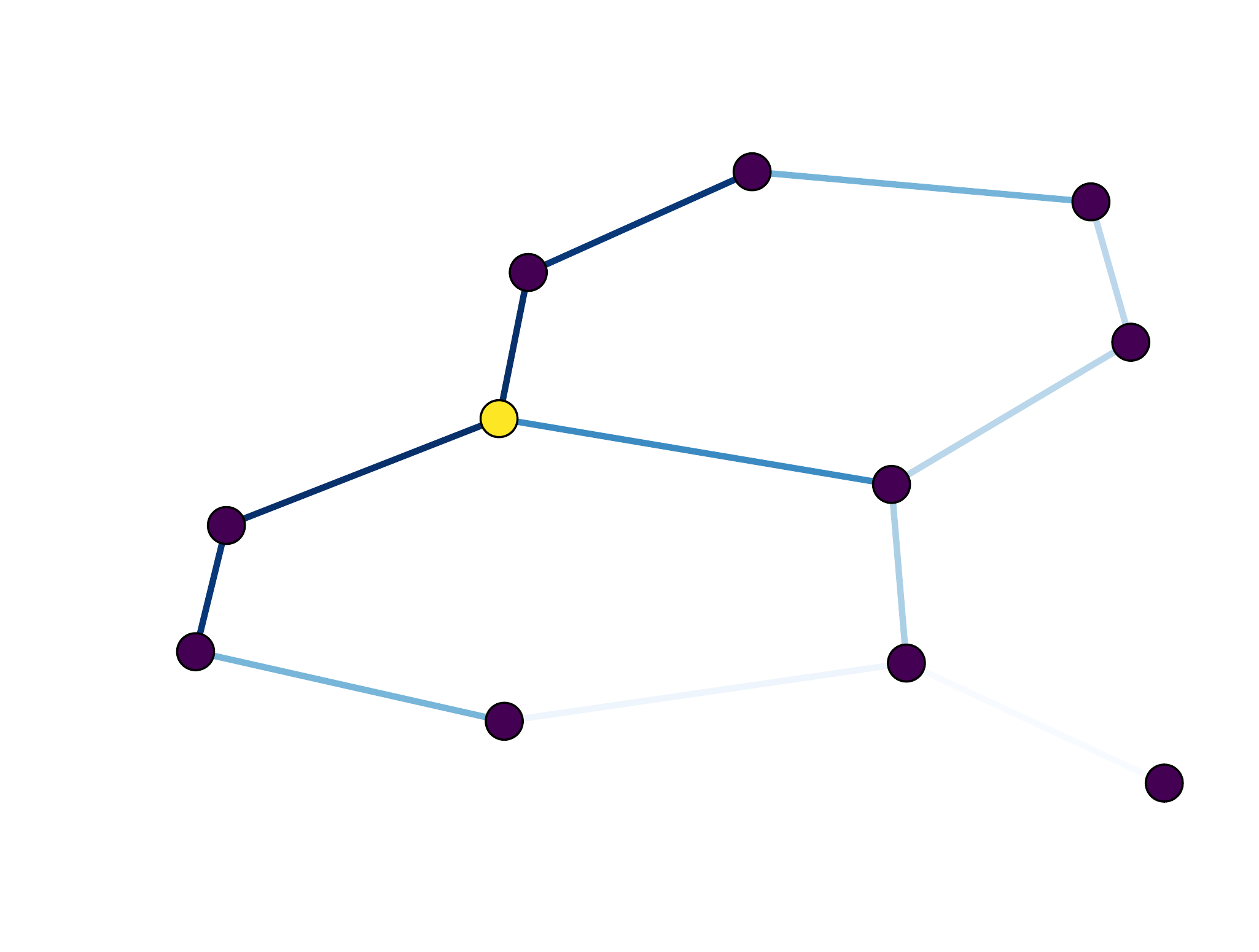}
\label{fig:sigmoid_m3}
}
\centering
\subfigure[\scriptsize{\emph{RelEx$_\text{Gumbel}$} }]{
\includegraphics[width=0.14\columnwidth]{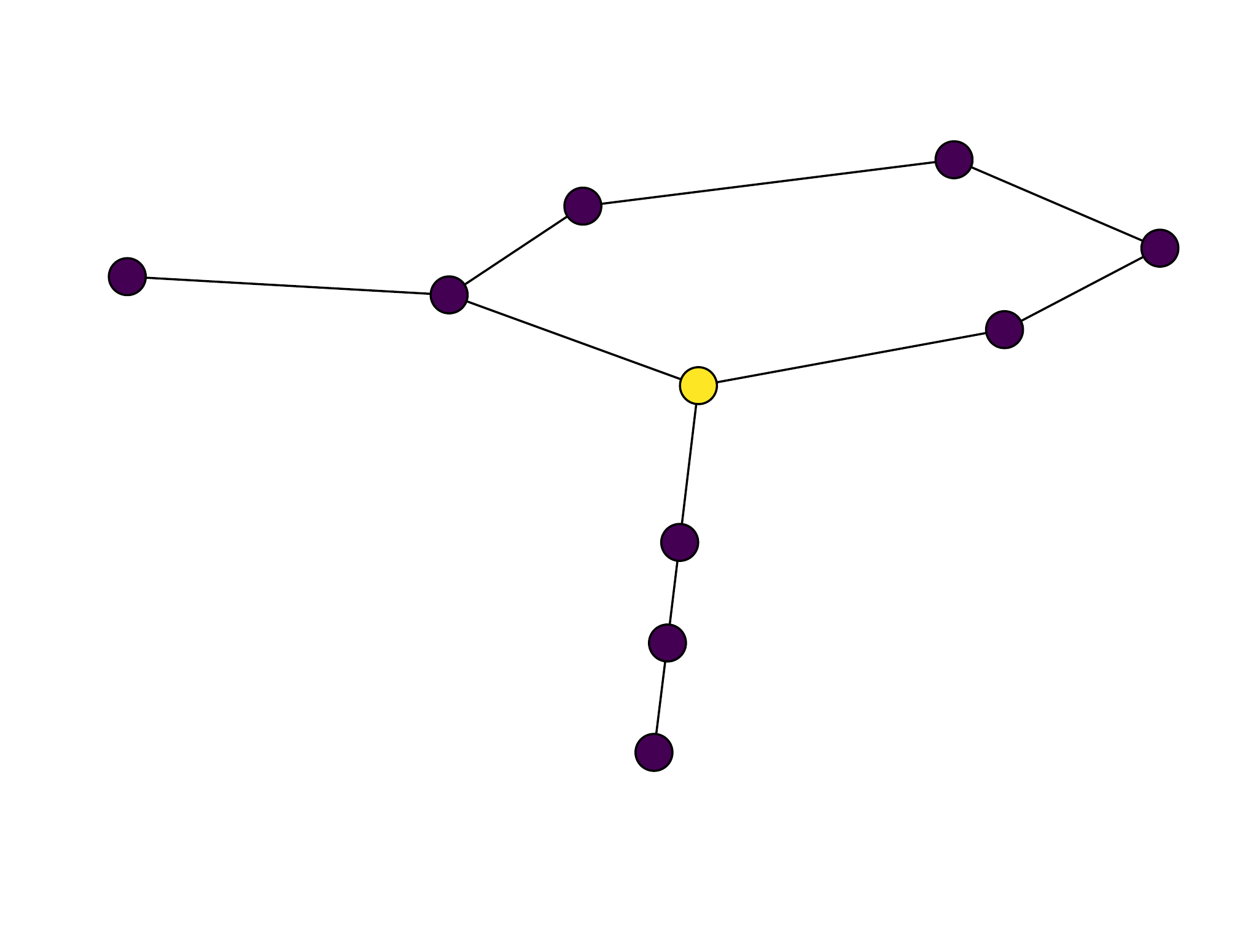}
\label{fig:gumbel_m2}
}
\caption{Relational explanations for a carbon atom.} 
\label{fig:mutag_carbon_1}

\centering
\subfigure[\scriptsize{Molecule}]{
\includegraphics[width=0.14\columnwidth]{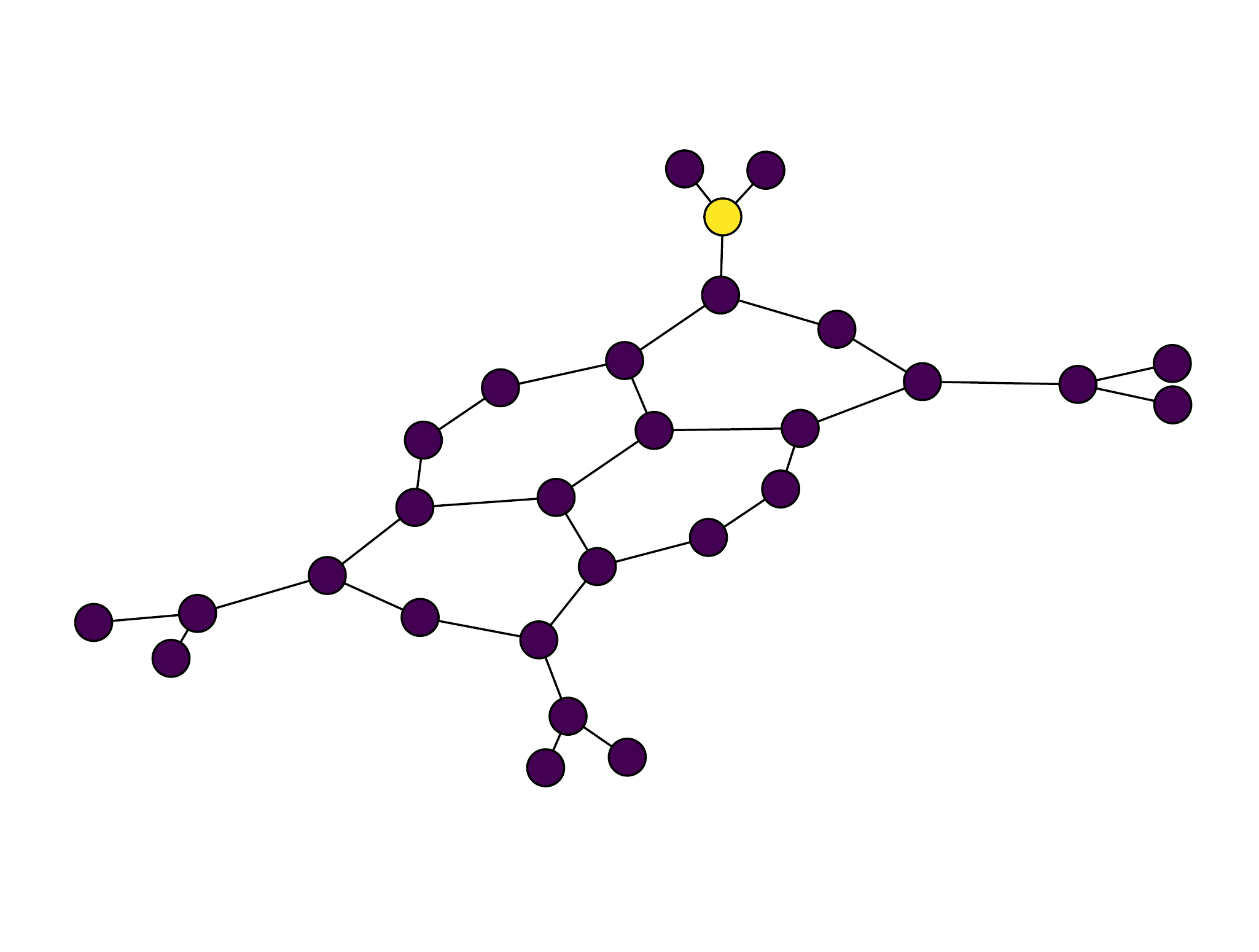}
\label{fig:molecule_m1}
}
\centering
\subfigure[\scriptsize{Right Reason}]{
\includegraphics[width=0.14\columnwidth]{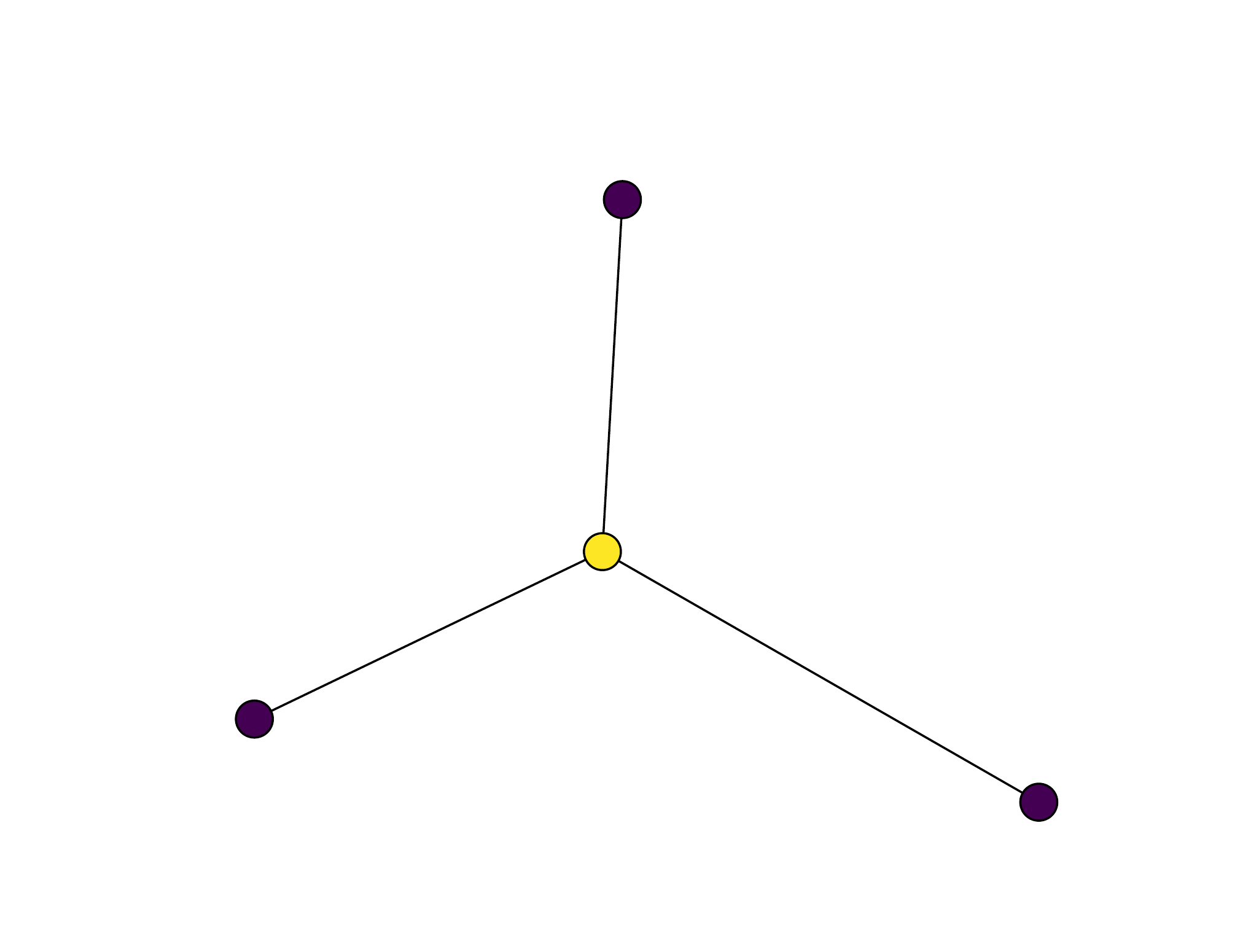}
\label{fig:rr_m3}
}
\centering
\subfigure[\scriptsize{Relational Anchors}]{
\includegraphics[width=0.14\columnwidth]{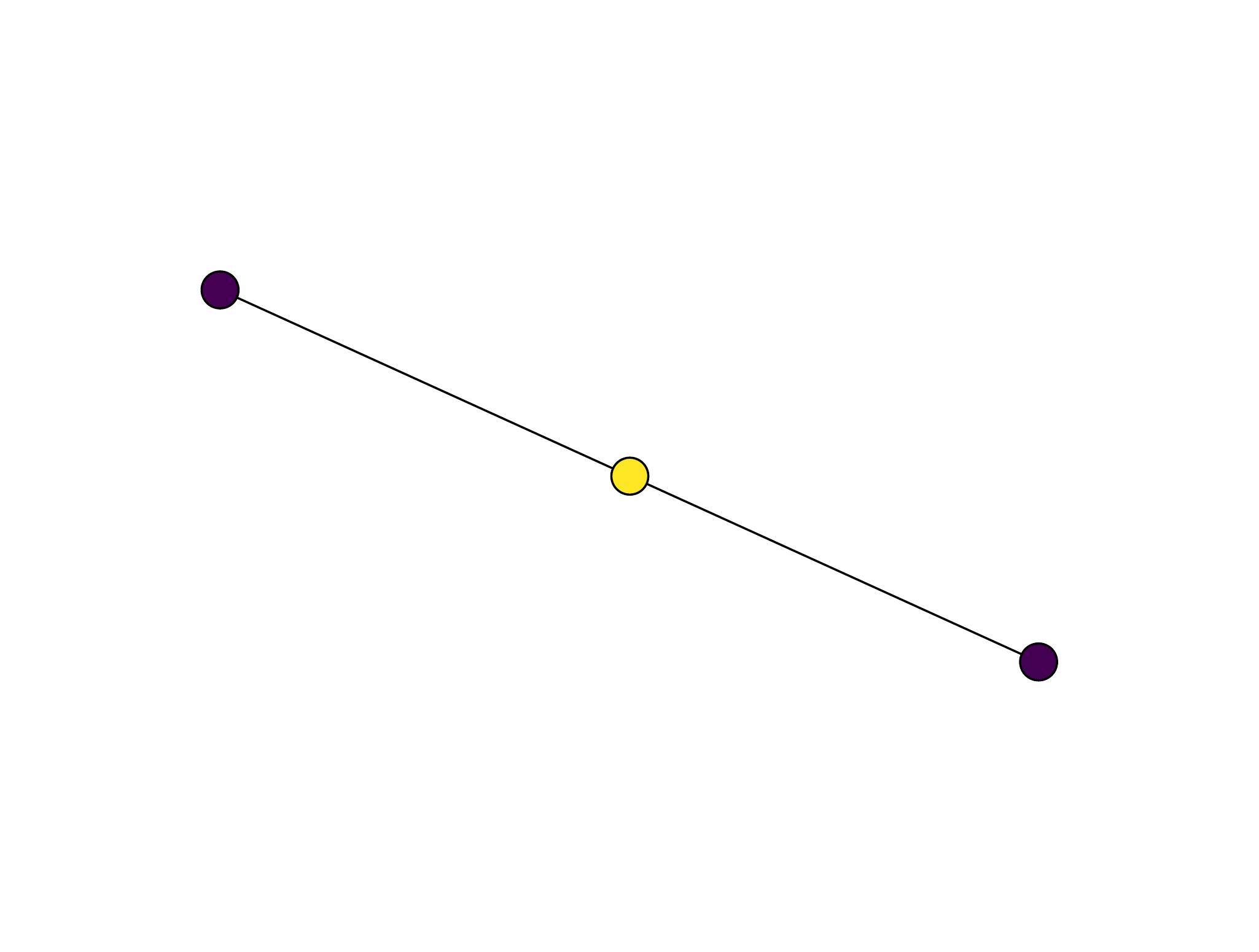}
\label{fig:anchor_m3}
}
\centering
\subfigure[\scriptsize{GNN-Explainer}]{
\includegraphics[width=0.14\columnwidth]{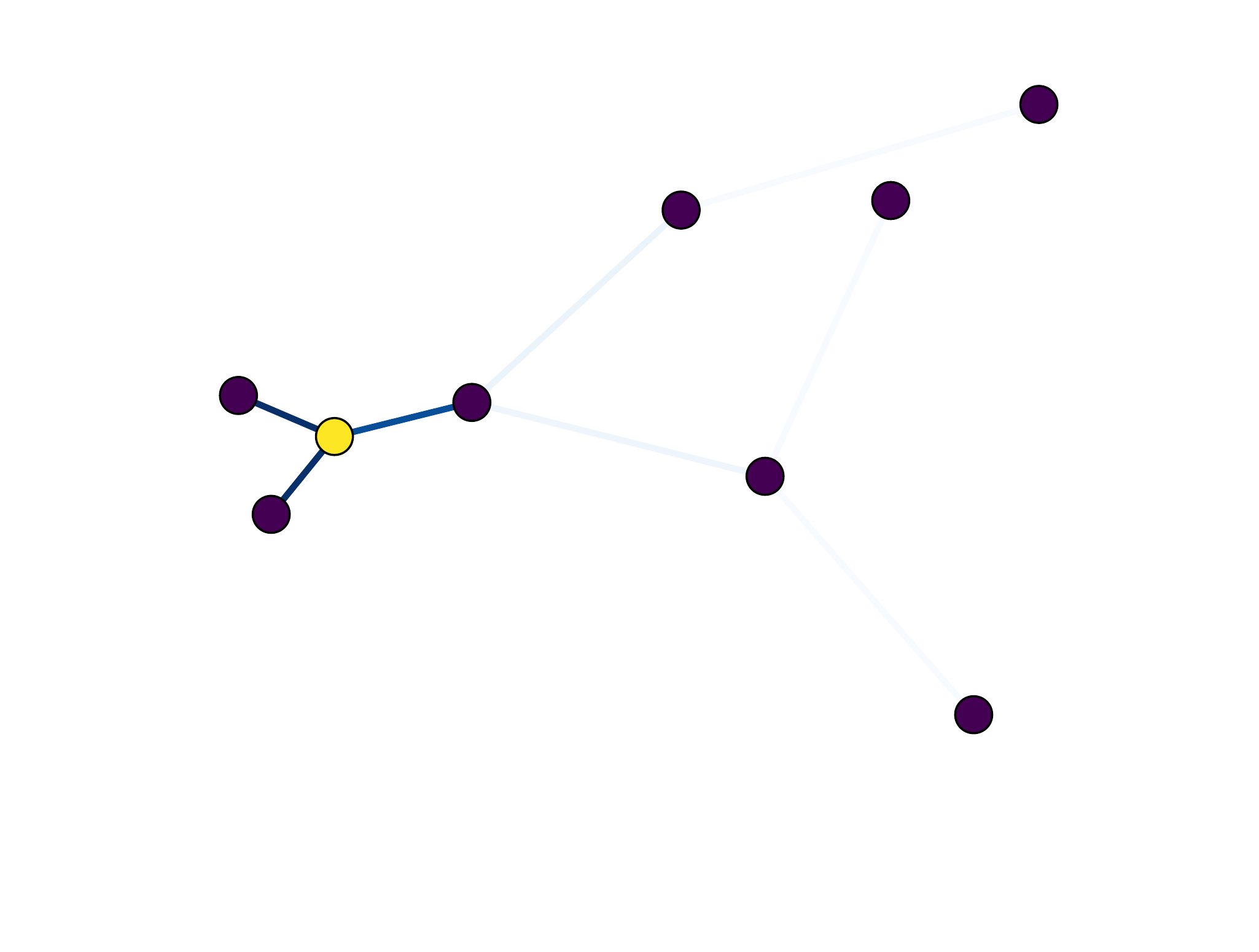}
\label{fig:gnnexplainer_m2}
}
\centering
\subfigure[\scriptsize{\emph{RelEx$_\text{Sigmoid}$}}]{
\includegraphics[width=0.14\columnwidth]{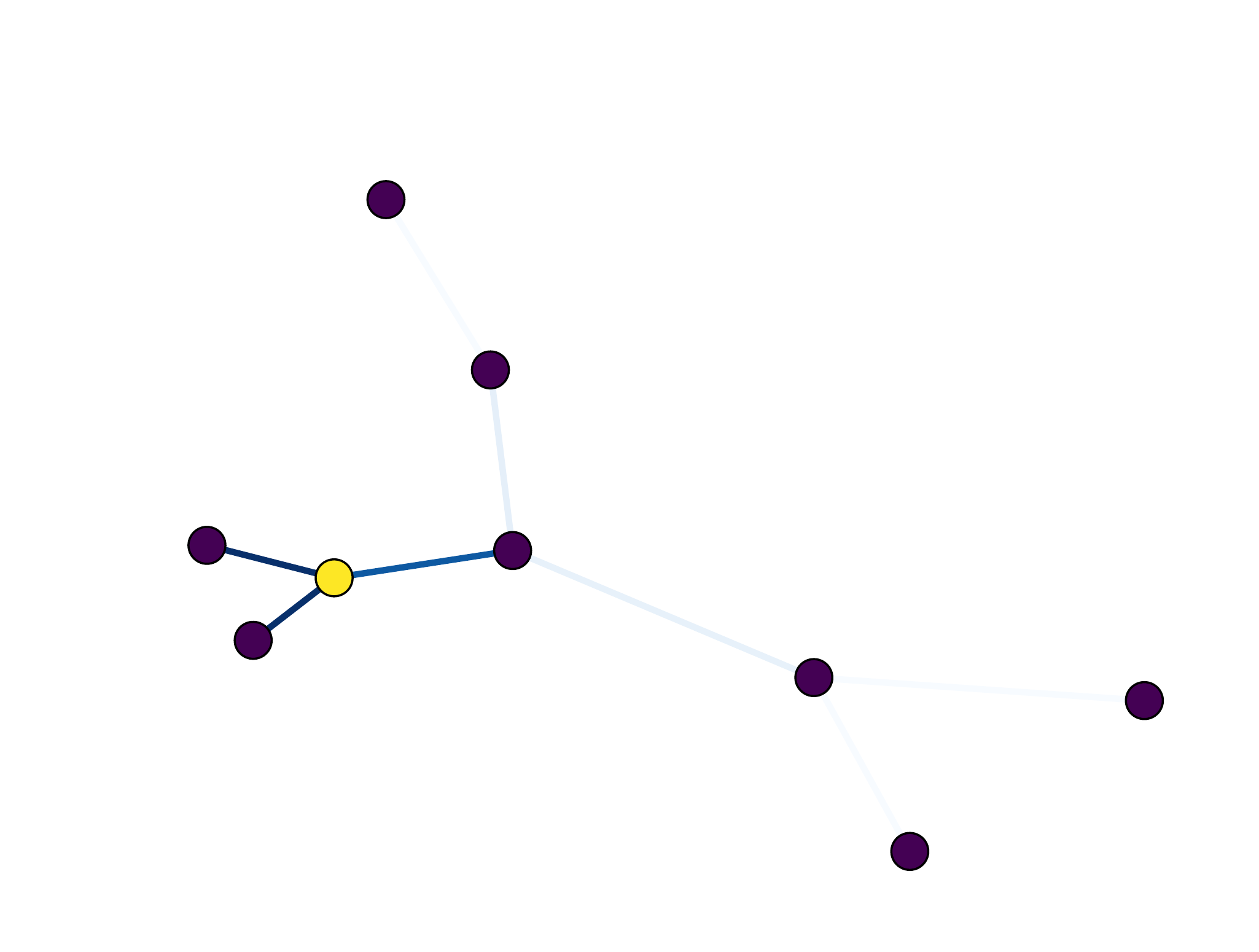}
\label{fig:sigmoid_m3}
}
\centering
\subfigure[\scriptsize{\emph{RelEx$_\text{Gumbel}$} }]{
\includegraphics[width=0.14\columnwidth]{figures/MUTAG/node90_class1/gumbelSoftmax.pdf}
\label{fig:gumbel_m2}
}
\caption{Explanation on nitrogen atom in $\text{-NO}_2$ structure. } 
\label{fig:mutag_nitrogen_1}
\end{figure}

Table \ref{table:MUTAG_results} shows the comparison results on infidelity, where we see that \emph{RelEx$_\text{Gumbel}$} and GNN-Explainer both obtain similar best results. We demonstrate the qualitative performance of the models in Figures \ref{fig:mutag_carbon_1}, and \ref{fig:mutag_nitrogen_1}. In all the figures, yellow nodes are our nodes of interest. To plot explanations from soft important values instead of finding the optimal threshold, we choose to capture the edge importance using the color of the edge, where a darker color signifies that the edge has a higher importance. We see that explanations from  GNN-Explainer and \emph{RelEx$_\text{Sigmoid}$} are plotted this way as they learn soft importance values for the edges in the relational explanation. In Figure \ref{fig:mutag_carbon_1}, we observe that the explanation for a carbon node learned by \emph{RelEx$_\text{Gumbel}$} finds the correct hexagonal ring structure and \emph{RelEx$_\text{Sigmoid}$} learns an explanation that contains two connected hexagonal rings, both of which capture the core relational structure (hexagonal ring) corresponding to the carbon node. Figure \ref{fig:mutag_nitrogen_1} shows explanations on a nitrogen node; all explainers except Relational Anchors are able to identify the correct -NO$_2$ topological structure.


\subsection{Diverse Explanations on Molecule Dataset}
\label{sec:diverseExperiment}
\begin{figure}[t]
\centering
\subfigure[\scriptsize{Molecule}]{
\includegraphics[width=0.14\columnwidth]{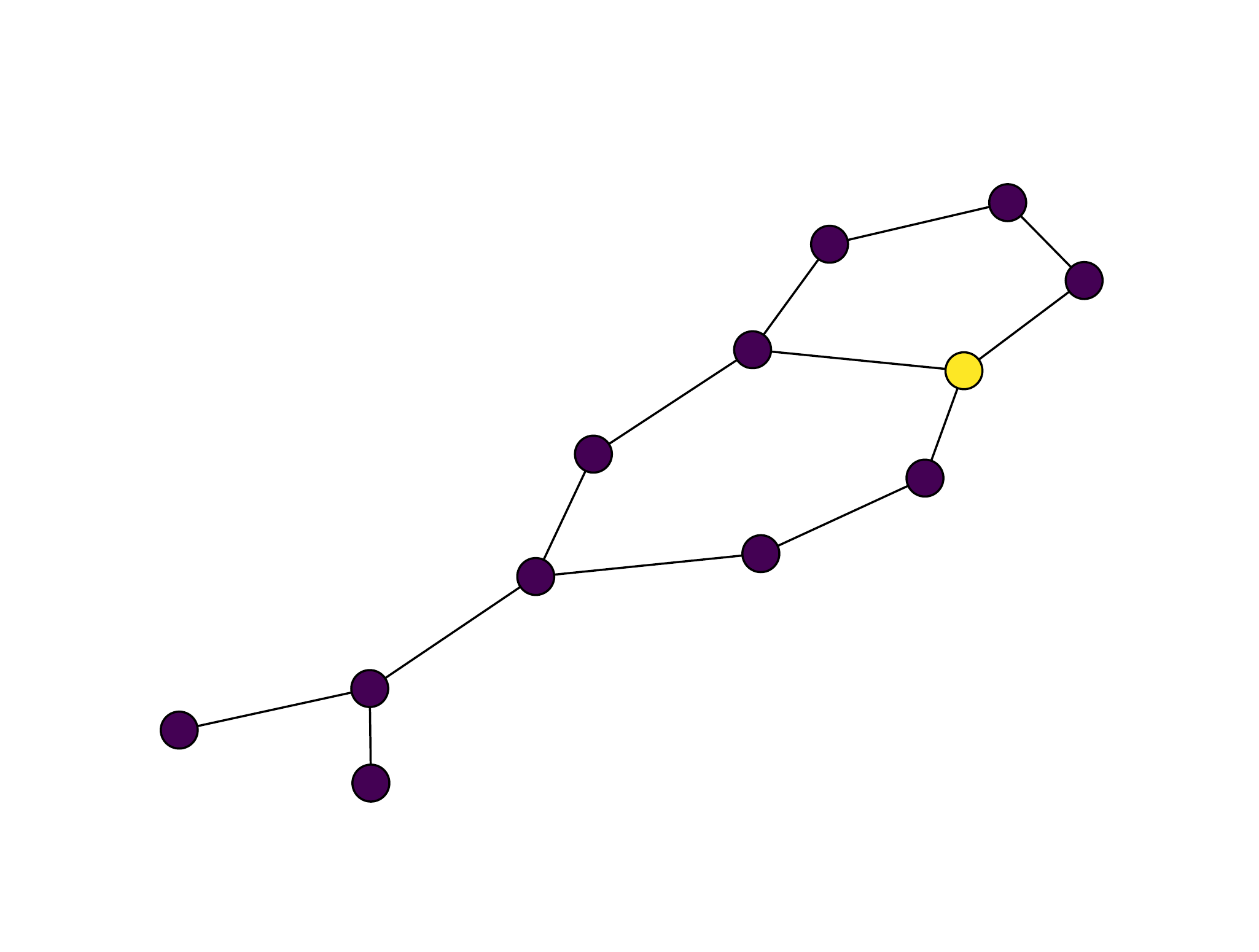}
\label{fig:d1m}
}
\centering
\subfigure[\scriptsize{Explanation 1}]{
\includegraphics[width=0.14\columnwidth]{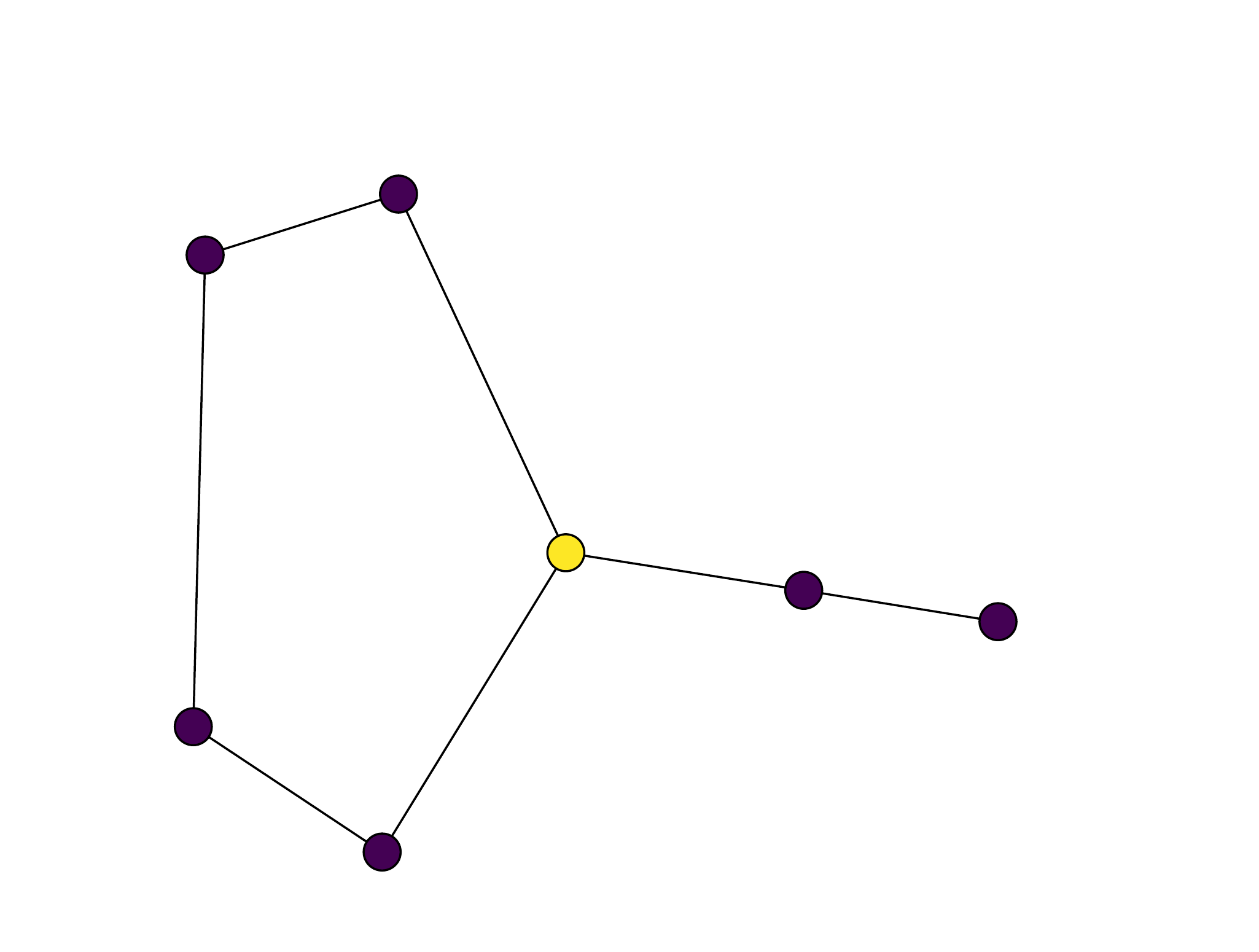}
\label{fig:d1e1}
}
\centering
\subfigure[\scriptsize{Explanation 2}]{
\includegraphics[width=0.14\columnwidth]{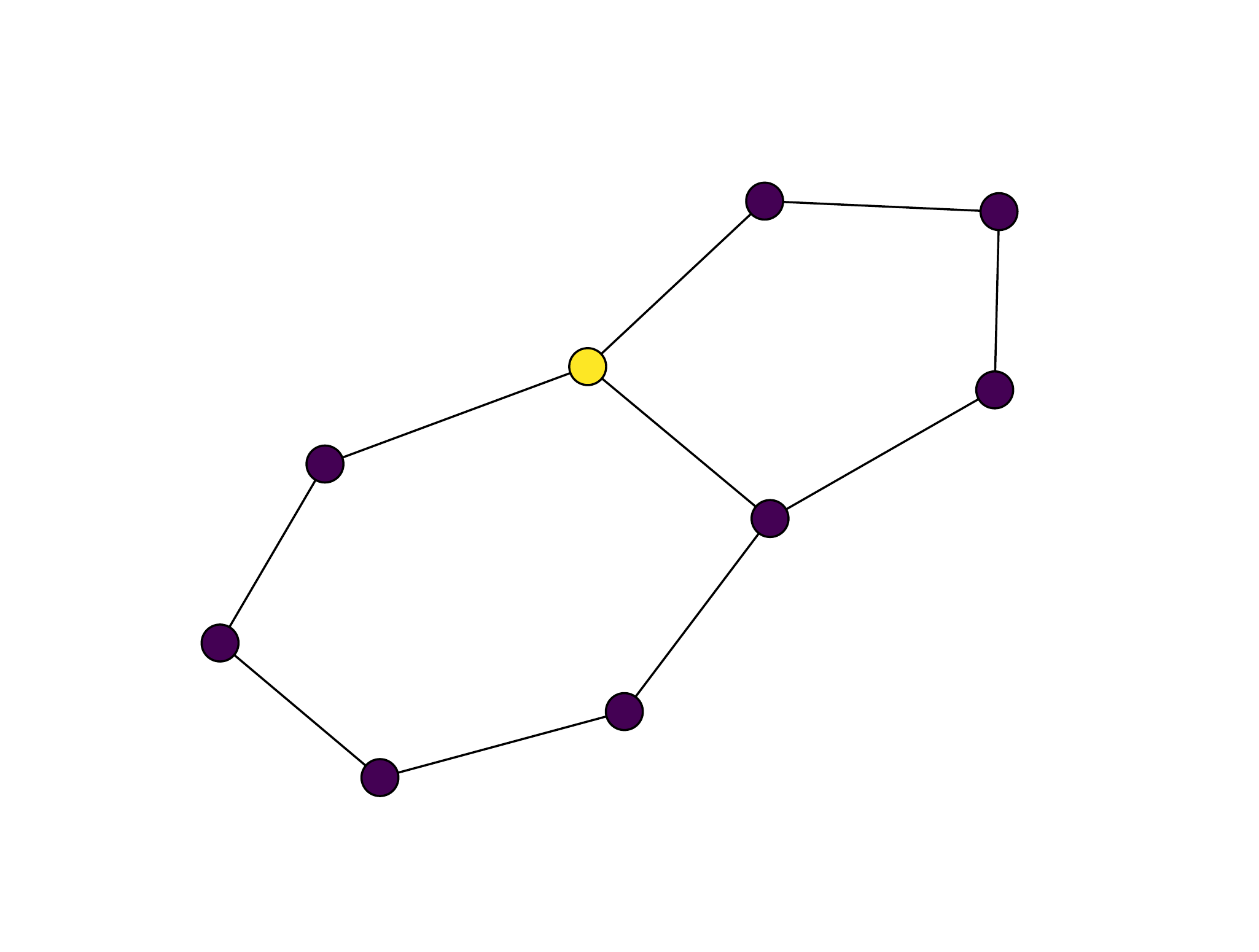}
\label{fig:d1e2}
}
\centering
\subfigure[\scriptsize{Molecule}]{
\includegraphics[width=0.14\columnwidth]{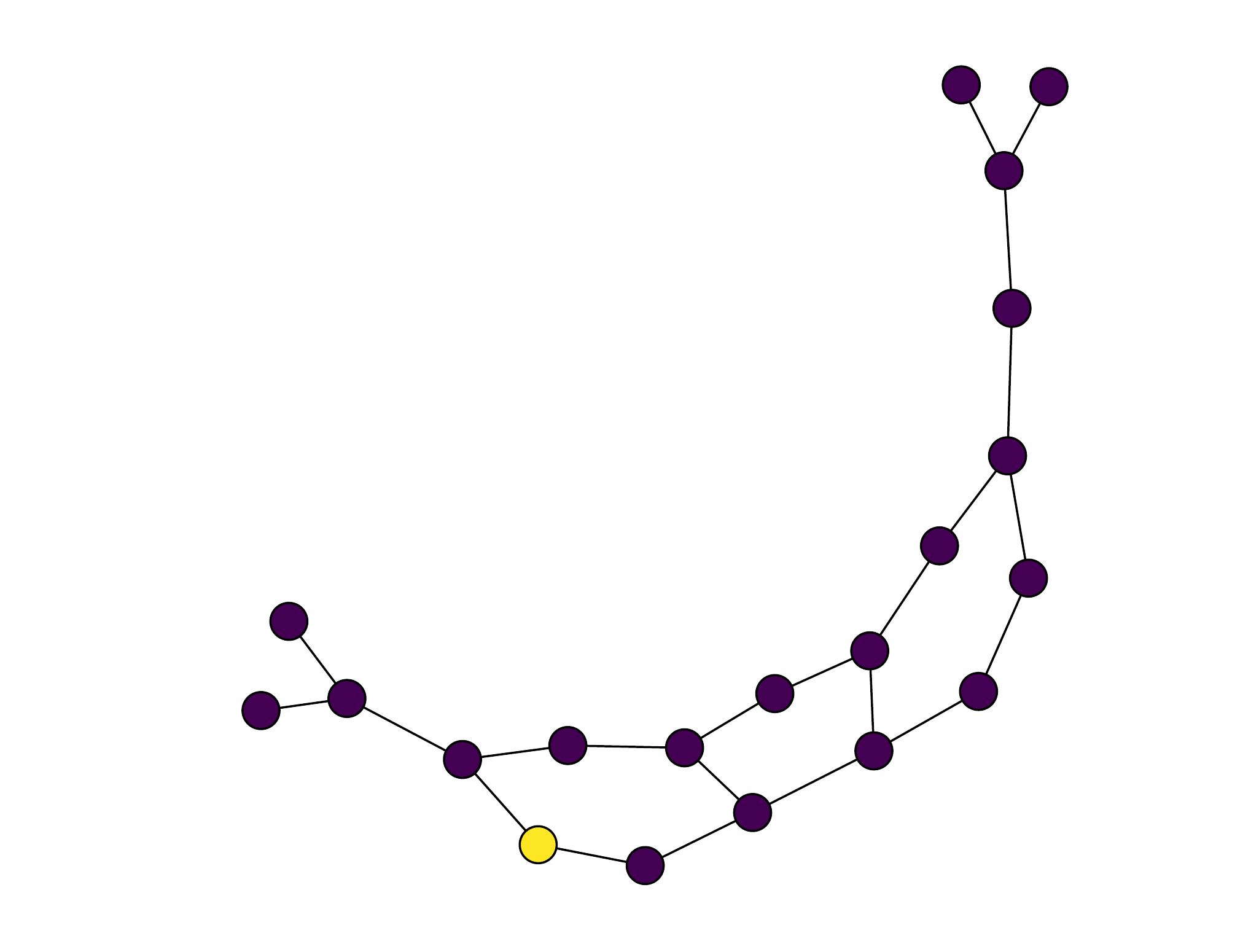}
\label{fig:d2m}
}
\centering
\subfigure[\scriptsize{Explanation 1}]{
\includegraphics[width=0.14\columnwidth]{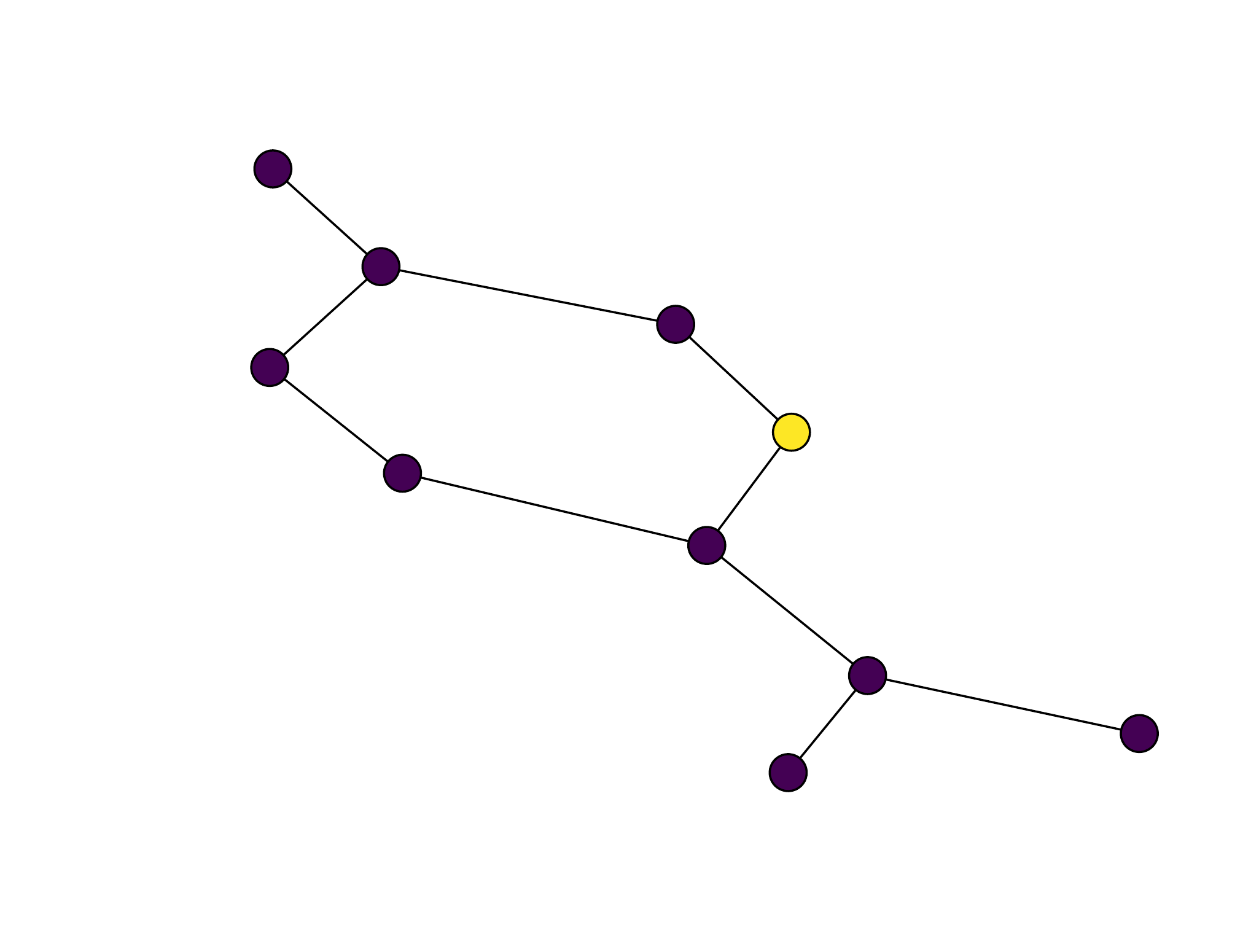}
\label{fig:d2e1}
}
\centering
\subfigure[\scriptsize{Explanation 2}]{
\includegraphics[width=0.14\columnwidth]{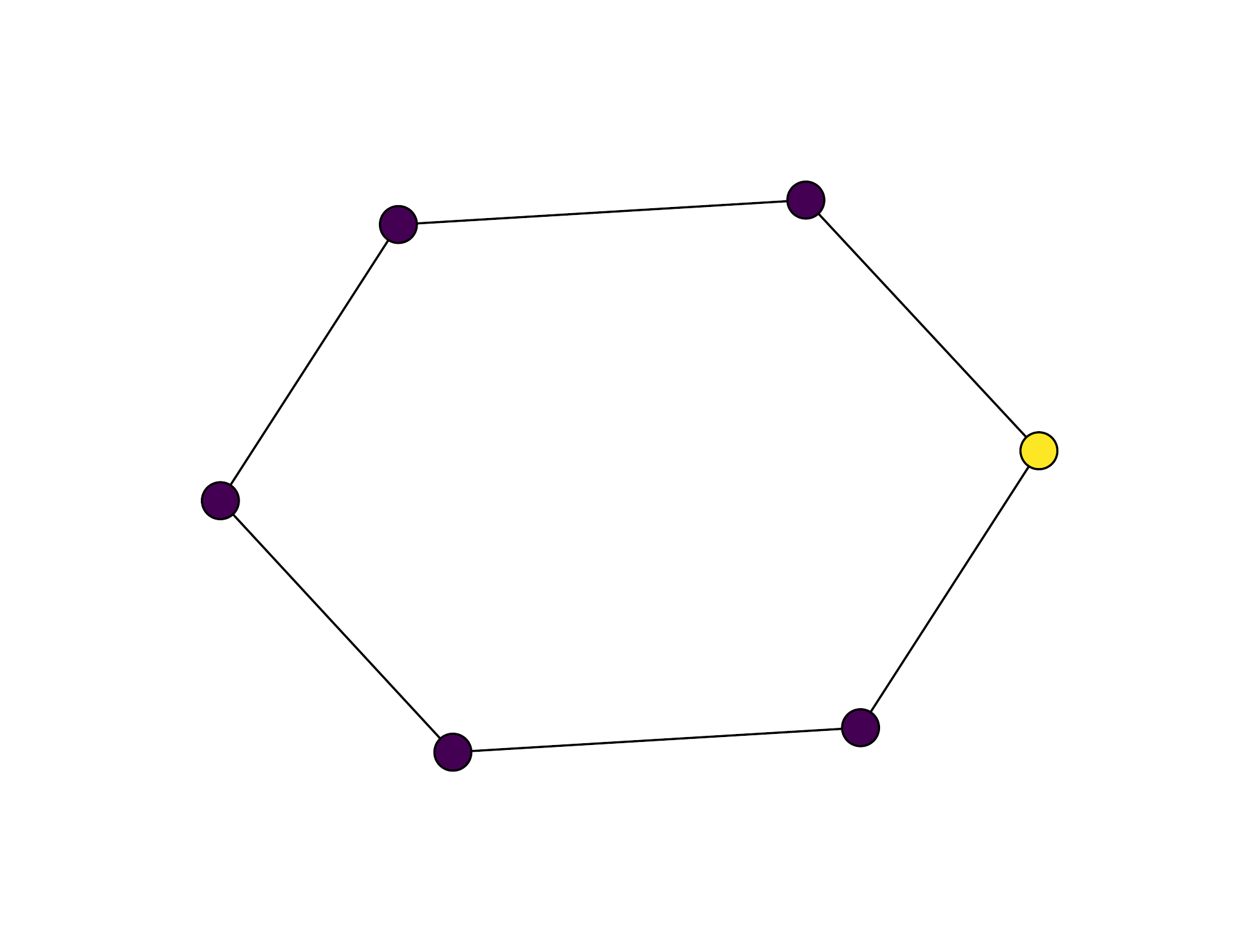}
\label{fig:d2e2}
}
\caption{Diverse explanations} 
\label{fig:diverse2}
\end{figure}

We train diverse explainations for each node of interest. Figure \ref{fig:diverse2} gives two example explanations learnt from the \emph{RelEx$_\text{Gumbel}$} based explainer, where yellow nodes are our nodes of interest. Figure \ref{fig:d1m} shows the molecule, and Figures \ref{fig:d1e1} and \ref{fig:d1e2} give two diverse explanations for the same node. In Figure \ref{fig:d1m}, we see that our node of interest is part of two ring structures, one of which is a pentagon and the other is a hexagon. The first explanation learns one pentagon ring structure, while the diverse second explanation finds both the ring structures. Though both are correct, the second explanation is more meaningful from the domain perspective as it gleans both the core relational structures that the node is part of. Similarly, in Figure \ref{fig:d2m}, even though both explanations are able to learn the core hexagonal structure responsible for the prediction, we see that the first explanation in Figure \ref{fig:d2e1} contains some noise, while the second diverse explanation in Figure \ref{fig:d2e2} excludes the noise and is more preferable. Thus, the ability of our approach to learn diverse explanations comes handy for learning multiple ``right'' explanations, among which some make more sense from a domain perspective. 

\section{Conclusion}

In this work, we developed a model-agnostic relational explainer, \emph{RelEx}, which has the ability to explain any black-box relational model. Through rigorous experimentation and comparison with state-of-the-art explainers, we demonstrated the quantitative and qualitative capability of \emph{RelEx} in explaining two different black-box relational models, GNNs, representing the deep graph neural network models, and HL-MRFs, representing statistical relational models, on two synthetic and one real-world graph datasets. The ability of \emph{RelEx} to learn diverse explanations further enhances its practical value and applicability in explaining domain-specific predictions. 


\bibliographystyle{nips}

\begin{thebibliography}{10}

\bibitem{srl2007}
Getoor, B. T.~L., ed.
\newblock \emph{Introduction to Statistical Relational Learning (Adaptive
  Computation and Machine Learning)}.
\newblock 2007.

\bibitem{scarselli2008graph}
Scarselli, F., M.~Gori, A.~C. Tsoi, et~al.
\newblock The graph neural network model.
\newblock \emph{IEEE Transactions on Neural Networks}, pages 61--80, 2008.

\bibitem{goodfellow2014explaining}
Goodfellow, I.~J., J.~Shlens, C.~Szegedy.
\newblock Explaining and harnessing adversarial examples.
\newblock \emph{arXiv}, 2014.

\bibitem{koh2017understanding}
Koh, P.~W., P.~Liang.
\newblock Understanding black-box predictions via influence functions.
\newblock In \emph{ICML}. 2017.

\bibitem{yeh2018representer}
Yeh, C.-K., J.~Kim, I.~E.-H. Yen, et~al.
\newblock Representer point selection for explaining deep neural networks.
\newblock In \emph{NeurIPS}. 2018.

\bibitem{sundararajan2017axiomatic}
Sundararajan, M., A.~Taly, Q.~Yan.
\newblock Axiomatic attribution for deep networks.
\newblock In \emph{ICML}. 2017.

\bibitem{smilkov2017smoothgrad}
Smilkov, D., N.~Thorat, B.~Kim, et~al.
\newblock Smoothgrad: removing noise by adding noise.
\newblock \emph{arXiv}, 2017.

\bibitem{ribeiro2016should}
Ribeiro, M.~T., S.~Singh, C.~Guestrin.
\newblock Why should i trust you?: Explaining the predictions of any
  classifier.
\newblock In \emph{SIGKDD}. 2016.

\bibitem{ribeiro2018anchors}
---.
\newblock Anchors: High-precision model-agnostic explanations.
\newblock In \emph{Thirty-Second AAAI Conference on Artificial Intelligence}.
  2018.

\bibitem{ying2019gnn}
Ying, R., D.~Bourgeois, J.~You, et~al.
\newblock Gnn explainer: A tool for post-hoc explanation of graph neural
  networks.
\newblock In \emph{NeurIPS}. 2019.

\bibitem{ross2017right}
Ross, A.~S., M.~C. Hughes, F.~Doshi-Velez.
\newblock Right for the right reasons: Training differentiable models by
  constraining their explanations.
\newblock \emph{arXiv}, 2017.

\bibitem{bach2017hinge}
Bach, S.~H., M.~Broecheler, B.~Huang, et~al.
\newblock Hinge-loss markov random fields and probabilistic soft logic.
\newblock \emph{JMLR}, pages 3846--3912, 2017.

\bibitem{kipf2017semi}
Kipf, T.~N., M.~Welling.
\newblock Semi-supervised classification with graph convolutional networks.
\newblock In \emph{ICLR}. 2017.

\bibitem{xu2018how}
Xu, K., W.~Hu, J.~Leskovec, et~al.
\newblock How powerful are graph neural networks?
\newblock In \emph{ICLR}. 2019.

\bibitem{velickovic2018graph}
Veli{\v{c}}kovi{\'{c}}, P., G.~Cucurull, A.~Casanova, et~al.
\newblock {Graph Attention Networks}.
\newblock \emph{ICLR}, 2018.

\bibitem{pmlr-v97-wu19e}
Wu, F., A.~Souza, T.~Zhang, et~al.
\newblock Simplifying graph convolutional networks.
\newblock In \emph{ICML}. 2019.

\bibitem{ma2019disentangled}
Ma, J., P.~Cui, K.~Kuang, et~al.
\newblock Disentangled graph convolutional networks.
\newblock In \emph{ICML}. 2019.

\bibitem{Fey/Lenssen/2019}
Fey, M., J.~E. Lenssen.
\newblock Fast graph representation learning with {PyTorch Geometric}.
\newblock In \emph{ICLR Workshop on Representation Learning on Graphs and
  Manifolds}. 2019.

\bibitem{he2016adaptive}
He, J., Y.~Zhang, Y.~Zhou, et~al.
\newblock Adaptive stochastic gradient descent on the grassmannian for robust
  low-rank subspace recovery.
\newblock \emph{IET Signal Processing}, pages 1000--1008, 2016.

\bibitem{zhang2019learning}
Zhang, Y., A.~Ramesh.
\newblock Learning interpretable relational structures of hinge-loss markov
  random fields.
\newblock In \emph{Proceedings of the 28th International Joint Conference on
  Artificial Intelligence}. AAAI Press, 2019.

\bibitem{dehmamy2019understanding}
Dehmamy, N., A.-L. Barab{\'a}si, R.~Yu.
\newblock Understanding the representation power of graph neural networks in
  learning graph topology.
\newblock In \emph{NeurIPS}. 2019.

\bibitem{jang2016categorical}
Jang, E., S.~Gu, B.~Poole.
\newblock Categorical reparameterization with gumbel-softmax.
\newblock \emph{arXiv}, 2016.

\bibitem{mothilal2020explaining}
Mothilal, R.~K., A.~Sharma, C.~Tan.
\newblock Explaining machine learning classifiers through diverse
  counterfactual explanations.
\newblock In \emph{FAT}. 2020.

\bibitem{yeh2019fidelity}
Yeh, C.-K., C.-Y. Hsieh, A.~Suggala, et~al.
\newblock On the (in) fidelity and sensitivity of explanations.
\newblock In \emph{NeurIPS}. 2019.

\bibitem{doi:10.1021/jm00106a046}
Debnath, A.~K., R.~L. Lopez~de Compadre, G.~Debnath, et~al.
\newblock Structure-activity relationship of mutagenic aromatic and
  heteroaromatic nitro compounds. correlation with molecular orbital energies
  and hydrophobicity.
\newblock \emph{Journal of Medicinal Chemistry}, pages 786--797, 1991.

\end{thebibliography}

\end{document}